\documentclass[10pt,journal,cspaper,compsoc]{IEEEtran}
\usepackage{amssymb,amsmath}
\usepackage{color}
\usepackage{multirow}
\usepackage{booktabs}
\newcommand{\tabincell}[2]{\begin{tabular}{@{}#1@{}}#2\end{tabular}}

\DeclareRobustCommand{\rchi}{{\mathpalette\irchi\relax}}
\newcommand{\irchi}[2]{\raisebox{\depth}{$#1\chi$}}

\ifCLASSOPTIONcompsoc
\else
\fi
\usepackage{graphicx}
\usepackage{subfigure}
\usepackage{rotating}
\usepackage{threeparttable}
\ifCLASSINFOpdf
\else
\fi
\begin{document}
%
\title{Multi-Directional Multi-Level Dual-Cross \\Patterns for Robust Face Recognition}
%
%
%
%
\author{Changxing~Ding,~\IEEEmembership{Student~Member,~IEEE},
        Jonghyun Choi,~\IEEEmembership{Student~Member,~IEEE},\\
        Dacheng~Tao,~\IEEEmembership{Fellow,~IEEE},
        and~Larry S.~Davis,~\IEEEmembership{Fellow,~IEEE}
\IEEEcompsocitemizethanks{
\IEEEcompsocthanksitem C. Ding and D. Tao are with the Centre for Quantum Computation $\&$ Intelligent Systems and the Faculty of Engineering and Information Technology,
University of Technology, Sydney, 81 Broadway Street, Ultimo, NSW 2007, Australia. E-mail: changxing.ding@student.uts.edu.au, dacheng.tao@uts.edu.au.
\IEEEcompsocthanksitem J. Choi and L.S.Davis are with the Institute for Advanced Computer Studies, University of Maryland, College Park, MD 20742.
E-mail: \{jhchoi,lsd\}@umiacs.umd.edu.}}



%
%


\markboth{IEEE TRANSACTIONS ON PATTERN ANALYSIS AND MACHINE INTELLIGENCE,~Vol.X, No.X, MONTH~20XX}%
{DING \MakeLowercase{\textit{et al.}}: Multi-Directional Multi-Level Dual-Cross Patterns for Robust Face Recognition}


%


\IEEEcompsoctitleabstractindextext{%
\begin{abstract}
To perform unconstrained face recognition robust to variations in illumination, pose and expression, this paper presents a new scheme to extract ``Multi-Directional Multi-Level Dual-Cross Patterns'' (MDML-DCPs) from face images. Specifically, the MDML-DCPs scheme exploits the first derivative of Gaussian operator to reduce the impact of differences in illumination and then computes the DCP feature at both the holistic and component levels. DCP is a novel face image descriptor inspired by the unique textural structure of human faces. It is computationally efficient and only doubles the cost of computing local binary patterns, yet is extremely robust to pose and expression variations. MDML-DCPs comprehensively yet efficiently encodes the invariant characteristics of a face image from multiple levels into patterns that are highly discriminative of inter-personal differences but robust to intra-personal variations.

Experimental results on the FERET, CAS-PERL-R1, FRGC 2.0, and LFW databases indicate that DCP outperforms the state-of-the-art local descriptors
(e.g. LBP, LTP, LPQ, POEM, tLBP, and LGXP) for both face identification and face verification tasks.
More impressively, the best performance is achieved on the challenging LFW and FRGC 2.0 databases by deploying MDML-DCPs in a simple recognition scheme.
\end{abstract}

}

\maketitle


%

\section{Introduction}
%
%

%
%
%
%

\IEEEPARstart{F}{ace} recognition has been an active area of research due to both the scientific challenge and its potential use in a wide range of practical applications.
Satisfactory performance has been achieved but often only in controlled environments.
More recently, there has been increased demand for recognition of unconstrained face images,
such as those collected from the internet~\cite{LFWTech} or captured by mobile devices and surveillance cameras~\cite{Ross2015report}.
However, recognition of unconstrained face images is a difficult problem due to degradation of face image quality and the wide variations of pose,
illumination, expression, and occlusion often encountered in images~\cite{ding2015comprehensive}.

A face recognition system usually consists of a face representation stage and a face matching stage.
For face matching, multi-class classifiers have been used for face identification,
such as the nearest neighbor (NN) classifier and the sparse representation classifier (SRC)~\cite{wright2009robust},
with two-class classifiers, such as support vector machine (SVM) and Bayesian analysis, being used for verification.
In face representation, good representations discriminate inter-personal differences while being robust to intra-personal variations.
Two major categories of face representation methods dominate recent research,
namely face image descriptor-based methods~\cite{ahonen2006face,cao2010face,xie2010fusing,chan2012multiscale}
and deep learning-based methods~\cite{taigman2014deepface,sun2014deep}.
The former has the advantage of ease of use, data independence, and robustness to real-life challenges such as illumination and expression differences.
Consequently, the design of effective face image descriptors is regarded as fundamental for face recognition.
Based on design methodology, we can group existing face image descriptors into two groups: hand-crafted descriptors and learning-based descriptors.

Most face image descriptors are hand-crafted, of which Local Binary Patterns (LBP) and Gabor wavelets are two representative methods;
Ahonen et al.~\cite{ahonen2006face} showed that the texture descriptor LBP is extremely effective for face recognition.
LBP works by encoding the gray-value differences between each central pixel and its neighboring pixels into binary codes;
the face image is then represented as the concatenated spatial histograms of the binary codes.
Many variants of LBP have been proposed.
For example, Local Ternary Patterns (LTP)~\cite{tan2010enhanced}
was proposed to enhance the robustness of LBP to noises.
Transition LBP (tLBP)~\cite{trefny2010extended} and Direction coded LBP (dLBP)~\cite{trefny2010extended}
were proposed to extract complementary information to LBP using novel encoding strategies.
On the other hand, Gabor wavelets aim to encode multi-scale and multi-orientation information of face images~\cite{zhang2007histogram,zhang2005local}.
LBP-like descriptors are therefore better at encoding fine-scale information while Gabor-based descriptors extract information on larger scales.
Other notable descriptors for face recognition include Local Phase Quantization (LPQ)~\cite{ahonen2008recognition}
and Patterns of Oriented Edge Magnitudes (POEM)~\cite{vu2012enhanced}.
In LPQ, the blur-invariant information in face images is encoded and the method shows good performance in unconstrained conditions~\cite{chan2012multiscale}.
POEM works by computing the LBP feature on orientated edge magnitudes.

Of the learning-based descriptors, LE~\cite{cao2010face}, Local Quantized Patterns (LQP)~\cite{ul2012visual}, and Discriminant Face Descriptor (DFD)~\cite{lei2014learning} have emerged in recent years, which rely on unsupervised or supervised learning techniques to optimize encoders. One important advantage of learning-based descriptors over hand-crafted descriptors is their greater diversity in sampling pattern shapes and the larger sampling size.

Despite the successful application of existing face image descriptors, the following three points are worth considering.
First, the textual characteristics of human faces have mostly been overlooked in the design of existing descriptors.
Second, it is generally prohibitive for hand-crafted descriptors to adopt a large sampling size due to the complication
of the resulting encoding scheme and large feature size (i.e., the number of histogram bins)~\cite{ul2012visual}.
However, a large sampling size is desirable since it provides better discriminative power,
as has been proved by learning-based descriptors;
it is therefore reasonable to ask whether it is genuinely impossible for hand-crafted descriptors to exploit large sampling sizes.
Third, the recently proposed descriptors achieve good performance but at the cost of using computationally expensive techniques
such as Gabor filtering and codebook learning.
It would therefore be desirable to obtain a face image descriptor with superior performance that retains a lower computational cost and feature size.

To address these three limitations of existing techniques, in this paper we present a novel face image descriptor named Dual-Cross Patterns (DCP).
Inspired by the unique textural structure of human faces, DCP encodes second-order discriminative information in the directions of major facial components:
the horizontal direction for eyes, eyebrows, and lips; the vertical direction for the nose; and the diagonal directions ($\pi/4$ and $3\pi/4$) for the end parts of facial components.
The sampling strategy we adopt samples twice as many pixels as LBP.
By appropriately grouping the sampled pixels from the perspective of maximum joint Shannon entropy, we keep the DCP feature size reasonable.
DCP is very efficient - only twice the computational cost of LBP.
Significantly better performance is achieved even when a sub-DCP (denoted herein as DCP-1 and DCP-2) of exactly the same time and memory costs as LBP is used.

A powerful face representation scheme is equally important for unconstrained face recognition as a good face image descriptor.
To improve the discrimination of face representations and even the more general image representations,
the usual approach is to fuse the information extracted by different descriptors~\cite{tan2010enhanced,wolf2011effective,xu2015multi}
and at different scales~\cite{chan2012multiscale,liu2009robust}.
To enhance robustness to pose variation,
Wright et al.~\cite{wright2009implicit} expanded image patch descriptors with their spatial locations and represented the face image as a histogram of quantized patch descriptors.
Ding et al.~\cite{ding2015multi} proposed to represent the face image using only the unoccluded facial texture that is automatically detected in the 3D pose normalized face image.
Recently, face representation has benefited from rapid progress in face alignment techniques~\cite{xiong2013supervised}.
For example, Chen et al.~\cite{chen2013blessing} built a high dimensional face representation by extracting LBP features from multi-scale image patches around dense facial feature points.
For other representative face representations, we direct readers to a recent survey~\cite{ding2015comprehensive}.

In this paper we propose a highly robust and discriminative face representation scheme called Multi-Directional Multi-Level Dual-Cross Patterns (MDML-DCPs).
Specifically, the MDML-DCPs scheme employs the first derivative of Gaussian operator to conduct multi-directional filtering to reduce the impact of differences in illumination.
DCP features are then computed at two levels: 1) holistic-level features incorporating facial contours and facial components and their configuration,
and 2) component-level features focusing on the description of a single facial component.
Thus, MDML-DCPs comprehensively encodes multi-level invariant characteristics of face images.

Both the DCP descriptor and the MDML-DCPs scheme are extensively evaluated on four large-scale databases: FERET~\cite{phillips2000feret}, CAS-PEAL-R1~\cite{gao2008cas},
FRGC 2.0~\cite{phillips2005overview}, and LFW~\cite{LFWTech}.
The proposed DCP descriptor consistently achieves superior performance for both face identification and face verification tasks.
More impressively, the proposed MDML-DCPs exploits only a single descriptor but achieves the best performance on two challenging unconstrained databases, FRGC 2.0 and LFW.
Besides, this paper provides a fair and systematic comparison between state-of-the-art facial descriptors,
which has been rarely performed in the face recognition field~\cite{akhtar2013face}.

The remainder of the paper is organized as follows: Section 2 details the DCP descriptor and the construction of the MDML-DCPs face representation scheme is discussed in Section 3.
Face recognition pipelines using MDML-DCPs are introduced in Section 4. Experimental results are presented in Section 5, leading to conclusions in Section 6.

\section{Dual-Cross Patterns}
The design of a face image descriptor consists of three main parts: image filtering, local sampling, and pattern encoding.
The implementation of image filtering is flexible: possible methods include Gabor wavelets~\cite{xie2010fusing},
Difference of Gaussian (DoG), or the recently proposed discriminative image filter~\cite{lei2014learning}.
In this paper, we focus on local sampling and pattern encoding, which are the core components of a face image descriptor.

\subsection{Local Sampling}
The essence of DCP is to perform local sampling and pattern encoding in the most informative directions contained within face images.
For face recognition, useful face image information consists of two parts: the configuration of facial components and the shape of each facial component.
The shape of facial components is, in fact, rather regular.
After geometric normalization of the face image, the central parts of several facial components,
i.e., the eyebrows, eyes, nose, and mouth, extend either horizontally or vertically,
while their ends converge in approximately diagonal directions ($\pi/4$ and $3\pi/4$).
In addition, wrinkles in the forehead lie flat, while those in the cheeks are either raised or inclined.

\begin{figure}
\centering
\includegraphics[width=2.3in]{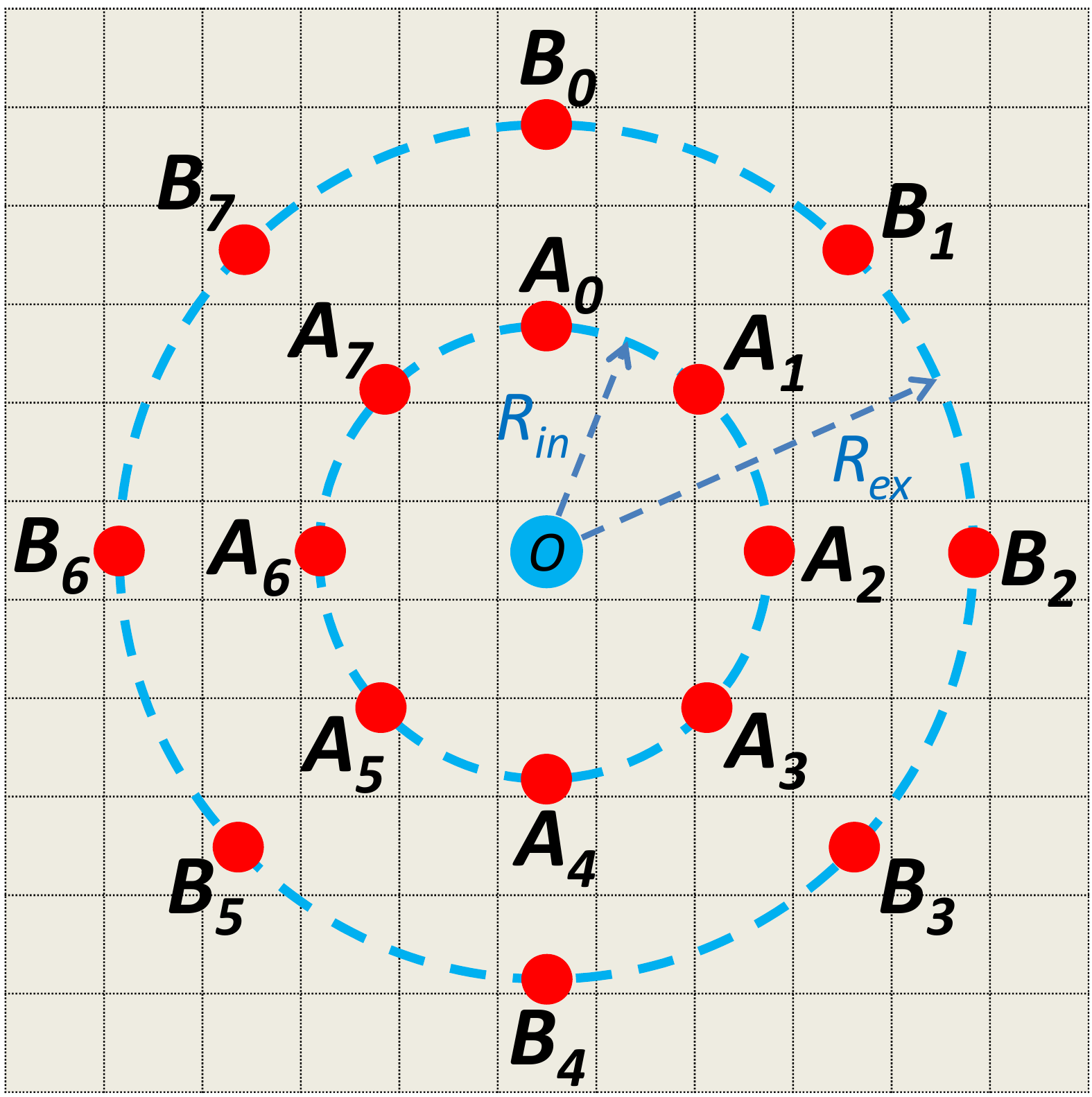}
\caption{Local sampling of Dual-Cross Patterns. Sixteen points are sampled around the central pixel $O$.
The sampled points ${A_0}$ to ${A_7}$ are uniformly spaced on an inner circle of radius $R_{in}$,
while ${B_0}$ to ${B_7}$ are evenly distributed on the exterior circle with radius $R_{ex}$.}
\label{fig:LocalSampling}
\end{figure}

Based on the above observations, local sampling of DCP is conducted as shown in Fig.~\ref{fig:LocalSampling}.
For each pixel $O$ in the image, we symmetrically sample in the local neighborhood in the $0$, $\pi/4$, $\pi/2$, $3\pi/4$, $\pi$, $5\pi/4$, $3\pi/2$, and $7\pi/4$ directions,
which are sufficient to summarize the extension directions of major facial textures. Two pixels are sampled in each direction.
The resulting sampled points are denoted as $\{{A_0},{B_0};{A_1},{B_1}; \cdots ;{A_7},{B_7}\}$.
As illustrated in Fig.~\ref{fig:LocalSampling}, ${A_0},{A_1}, \cdots ,{A_7}$ are uniformly spaced on an inner circle of radius $R_{in}$,
while ${B_0},{B_1} \cdots ,{B_7}$ are evenly distributed on the exterior circle with radius $R_{ex}$.

\subsection{Pattern Encoding}
Encoding of the sampled points is realized in two steps. First, textural information in each of the eight directions is independently encoded.
Second, patterns in all eight directions are combined to form the DCP codes.

To quantize the textural information in each sampling direction, we assign each a unique decimal number:
\begin{equation}
DC{P_i} = S\left( {{I_{{A_i}}} - {I_O}} \right) \times 2 + S\left( {{I_{{B_i}}} - {I_{{A_i}}}} \right),{\rm{  0}} \le i \le 7,
\end{equation}
where
\begin{equation}
S\left( x \right) = \left\{ \begin{array}{l}
 1,{\rm{ }}x \ge 0 \\
 0,{\rm{ }}x < 0\\
 \end{array} \right.,
\end{equation}
and ${{I_O}}$, ${{I_{{A_i}}}}$, and ${{I_{{B_i}}}}$ are the gray value of points $O$, ${A_i}$, and ${B_i}$, respectively.
Therefore, four patterns are defined to encode the second-order statistics in each direction and each of the four patterns denotes one type of textural structure.

By simultaneously considering all eight directions, the total number of DCP codes is ${4^8} = 65536$.
This number is too large for practical face recognition applications;
therefore, we adopt the following strategy.
The eight directions are grouped into two subsets and each subset is further formulated as an encoder.
In this way, the total number of local patterns is reduced to ${4^4} \times 2 = 512$, which is computationally efficient.
Although this strategy results in information loss, compactness and robustness of the descriptor are promoted.
In the following subsection, we define the optimal grouping mode.

\subsection{Dual-Cross Grouping}
The grouping strategy introduced in the previous subsection produces $\frac{1}{2}\binom{8}{4}=35$ combinations in total to partition all eight directions.
To minimize information loss, we look for the optimal combination from the perspective of maximum joint Shannon entropy.

With the above analysis, $DC{P_i}\left( {0 \le i \le 7} \right)$ are discrete variables with four possible values: 0, 1, 2, and 3.
Without loss of generality, the joint Shannon entropy for the subset $\left\{ {DC{P_0},{\rm{ }}DC{P_1},{\rm{ }}DC{P_2},{\rm{ }}DC{P_3}} \right\}$ is represented as
\begin{equation}
\label{eq:jointEntropy}
\begin{array}{l}
H\left( {DC{P_0},DC{P_1},DC{P_2},DC{P_3}} \right) \\
=  - \sum\limits_{dc{p_0}}  \cdots  \sum\limits_{dc{p_3}} {P\left( {dc{p_0}, \cdots ,dc{p_3}} \right)   {{\log }_2}P\left( {dc{p_0}, \cdots ,dc{p_3}} \right)},  \\
\end{array}
\end{equation}
where $dc{p_0}$, $dc{p_1}$, $dc{p_2}$, and $dc{p_3}$ are particular values of $DC{P_0}$, $DC{P_1}$, $DC{P_2}$, and $DC{P_3}$, respectively.
And ${P\left( {dc{p_0}, \cdots ,dc{p_3}} \right)}$ is the probability of these values occurring simultaneously.
The maximum joint Shannon entropy of the four variables is achieved when they are statistically independent.

\begin{figure*}
\centering
\includegraphics[width=6in]{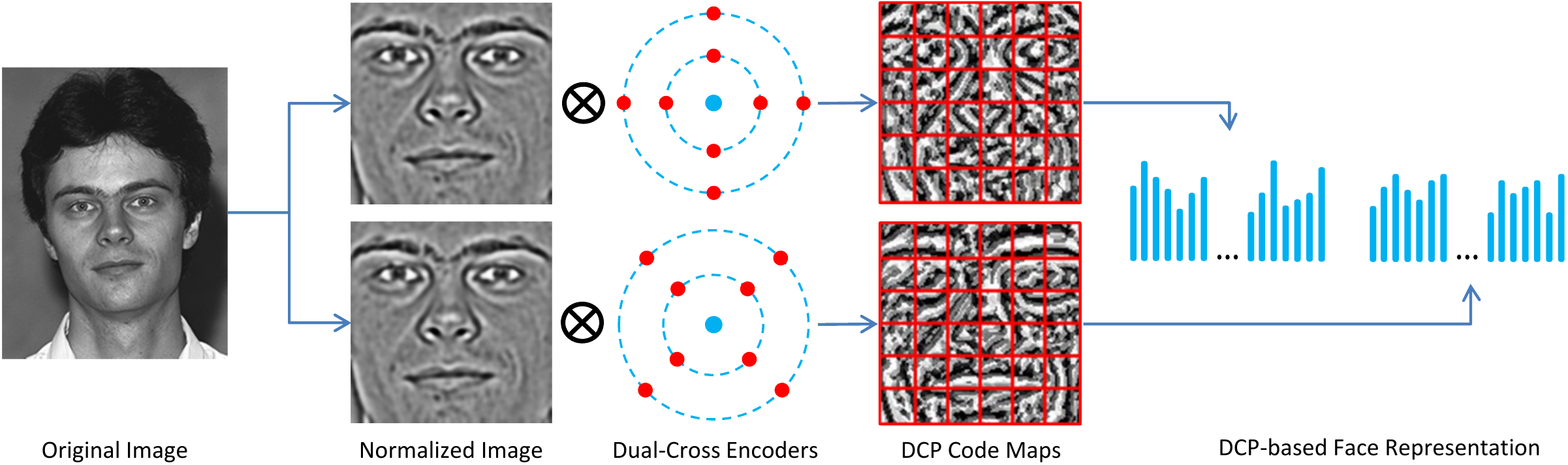}
\caption{Face representation using Dual-Cross Patterns. The normalized face image is encoded by the two cross encoders, respectively. Concatenation of the regional DCP code histograms forms the DCP-based face representation.}
\label{fig:DcpFlowChart}
\end{figure*}

In real images, the more sparsely the pixels are scattered, the more independent they are.
Therefore, the maximum joint Shannon entropy for each subset is achieved when the distance between the sampled points is at its maximum.
As a result, we define $\left\{ {DC{P_0},{\rm{ }}DC{P_2},{\rm{ }}DC{P_4},{\rm{ }}DC{P_6}} \right\}$ as the first subset
and $\left\{ {DC{P_1},{\rm{ }}DC{P_3},{\rm{ }}DC{P_5},{\rm{ }}DC{P_7}} \right\}$ as the second subset.
The resulting two subsets are illustrated in Fig.~\ref{fig:DcpFlowChart}.
Since each of the two subsets constructs the shape of a cross, the proposed descriptor is named Dual-Cross Patterns.
In Section 5.1, we empirically validate that dual-cross grouping achieves the maximum joint Shannon entropy among all grouping modes on the FERET database.

\subsection{DCP Face Image Descriptor}
We name the two cross encoders DCP-1 and DCP-2, respectively. The codes produced by the two encoders at each pixel $O$ are represented as
\begin{equation}
DCP\textrm{-}1 = \sum\limits_{i = 0}^3 {DC{P_{2i}} \times {4^i}},
\end{equation}
\begin{equation}
DCP\textrm{-}2 = \sum\limits_{i = 0}^3 {DC{P_{2i + 1}} \times {4^i}}.
\end{equation}

The DCP descriptor for each pixel $O$ in an image is the concatenation of the two codes generated by the two cross encoders:
\begin{equation}
DCP = \left\{ {\sum\limits_{i = 0}^3 {DC{P_{2i}} \times {4^i},{\rm{ }}\sum\limits_{i = 0}^3 {DC{P_{2i + 1}} \times {4^i}} } } \right\}.
\end{equation}

After encoding each pixel in the face image using the dual-cross encoders, two code maps are produced that are respectively divided into a grid of non-overlapping regions.
Histograms of DCP codes are computed in each region and all histograms are concatenated to form the holistic face representation.
The overall framework of the above face representation approach is illustrated in Fig.~\ref{fig:DcpFlowChart}.
This face representation can be directly used to measure the similarity between a pair of face images using metrics such as the chi-squared distance or histogram intersection.
The computation of the DCP descriptor, which doubles the feature size of LBP, is very efficient by only doubling the time cost of LBP.

We notice that two recently proposed descriptors DFD~\cite{lei2014learning} and Center Symmetric-Pairs of Pixels (CCS-POP)~\cite{choi2012complementary}
adopt similar sampling modes to DCP.
However, they are essentially different from DCP.
First, there is no clear motivation for the two descriptors that why such a sampling mode is suitable for face images. Second, the pattern encoding strategies are
different: both the two descriptors rely on learning algorithms to handle the large sampling size problem mentioned in Section 1.

\section{Multi-Directional Multi-Level Dual-Cross Patterns}
We now present a face representation scheme based on DCP named Multi-Directional Multi-Level Dual-Cross Patterns (MDML-DCPs)
to explicitly handle the challenges encountered in unconstrained face recognition.

\subsection{The MDML-DCPs Scheme}
A major difficulty in unconstrained face recognition is that many factors produce significant intra-personal differences in the appearance of face images,
in particular variations in illumination, image blur, occlusion, and pose and expression changes.
We mitigate the influence of these factors using multi-directional gradient filtering and multi-level face representation.

In MDML-DCPs, the first derivative of Gaussian operator (FDG) is exploited to convert a gray-scale face image into multi-directional gradient images that are more robust to variations in illumination. The FDG gradient filter of orientation $\theta$ can be expressed as follows:
\begin{equation}
FDG\left( \theta  \right) = \frac{{\partial G}}{{\partial \textbf{n}}} = \textbf{n}  \cdot \nabla G,
\end{equation}
where $\textbf{n}=(cos\theta, sin\theta)$ is the normal vector standing for the filtering direction and $G = \exp \left( { - \frac{{{x^2} + {y^2}}}{{{\sigma ^2}}}} \right)$ is a two-dimensional Gaussian filter. The application of FDG is inspired by the classical work of Canny~\cite{canny1986computational}, which proved that FDG is the optimal gradient filter according to three criteria, namely signal-to-noise ratio (SNR) maximization, edge location accuracy preservation, and single response to single edge. These three criteria are also relevant for face recognition, where it is desirable to enhance the facial textural information while suppressing noise. FDG significantly saves computational cost compared to other gradient-like filters, such as Gabor wavelets~\cite{xie2010fusing}. We denote the concatenation of DCP descriptors extracted from the FDG-filtered images as MD-DCPs.

To build pose-robust face representation, MDML-DCPs normalizes the face image by two geometric rectifications based on a similarity transformation and an affine transformation, respectively.
The similarity transformation retains the original information of facial contours and facial components and their configuration. The affine transformation reduces differences in intra-personal
appearance caused by pose variation.

\begin{figure*}
\centering
\includegraphics[width=5.4in]{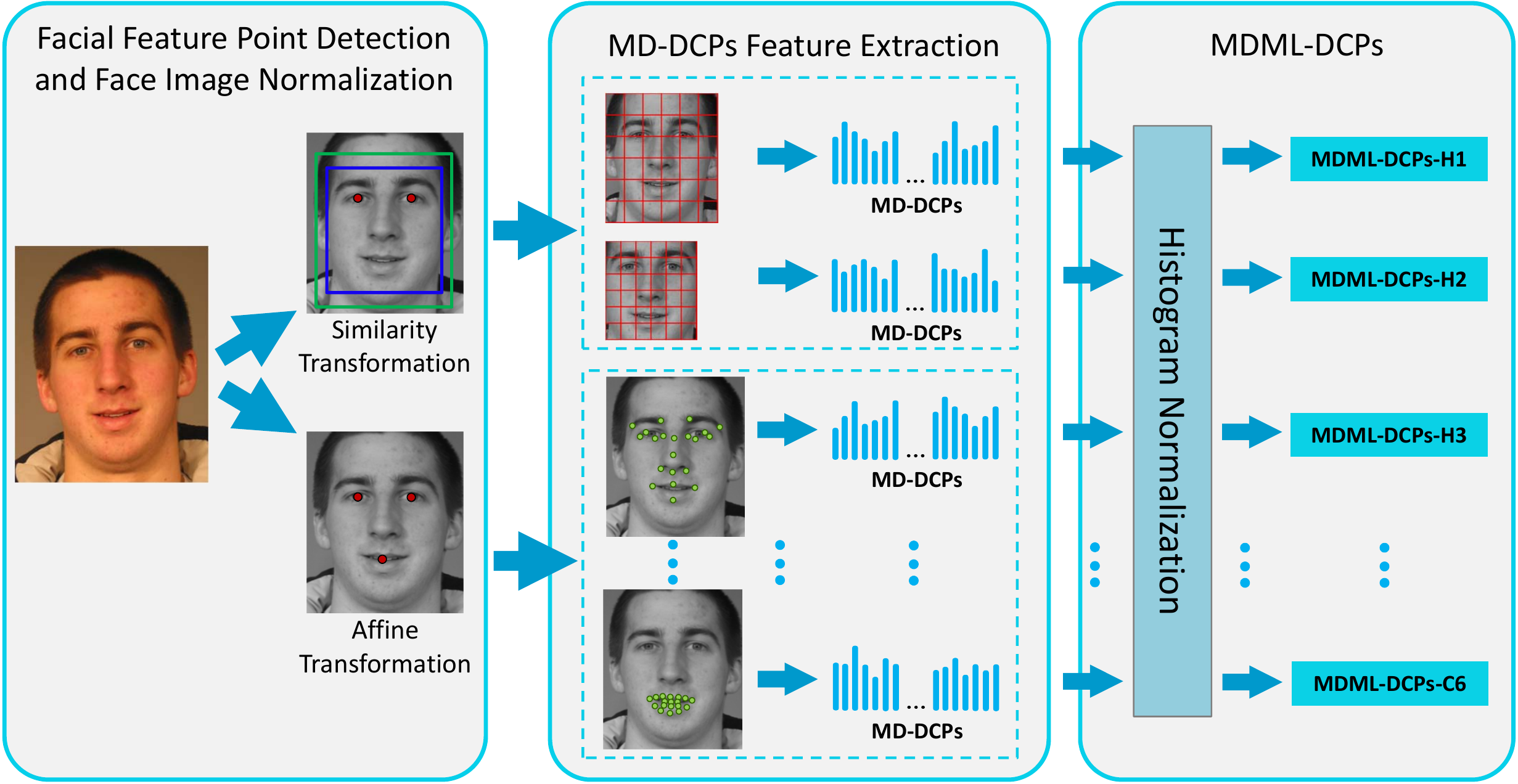}
\caption{Framework of the MDML-DCPs face representation scheme.
MDML-DCPs-H1 and MDML-DCPs-H2 are extracted from the rectified image by similarity transformation. MDML-DCPs-H3, MDML-DCPs-C1 to C6 are extracted from the affine-transformed image.
The MDML-DCPs face representation is the set of the above nine feature vectors.}
\label{fig:MDMLDcpsFramework}
\end{figure*}

\begin{figure}
\centering
\includegraphics[width=3in]{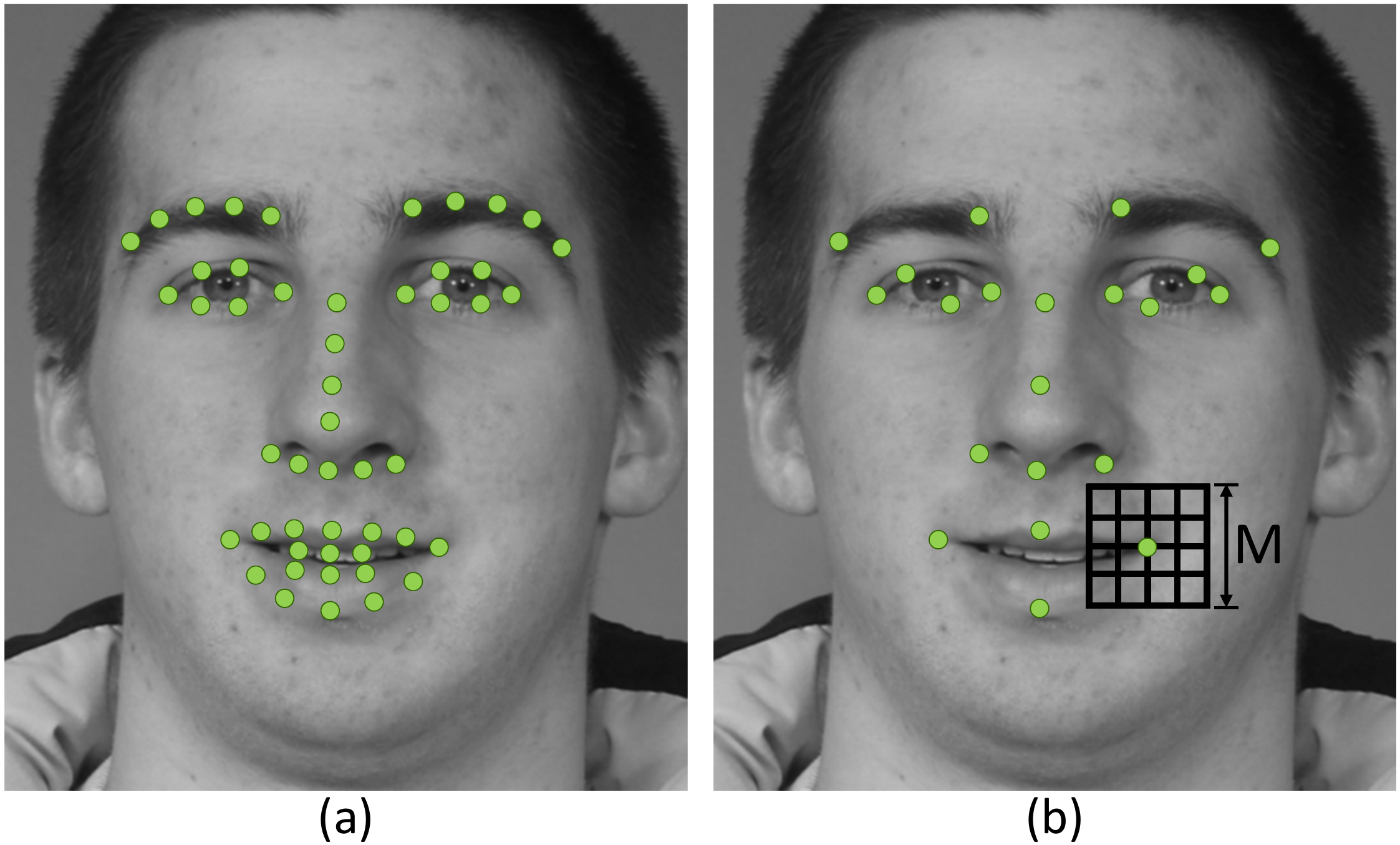}
\caption{(a) The 49 facial feature points detected by the face alignment algorithm.
(b) MDML-DCPs-H3 employs 21 facial feature points over all facial components.
MDML-DCPs-C1 to C6 respectively select 10 facial feature points on both eyebrows, 12 points on both eyes, 11 points on the left eye and left eyebrow,
11 points on the right eye and right eyebrow, 9 points on nose, and 18 points on mouth.
Around each facial feature point, MD-DCPs are extracted from $J\times J$ (in this figure, $J=4$) non-overlapping regions within the patch of size $M\times M$ pixels.
}
\label{fig:MDMLDcpsSixComponent}
\end{figure}

MDML-DCPs combines both holistic-level and component-level features, which are computed on the normalized images by the two transformations. Holistic-level features capture comprehensive information on both facial components and facial contour. However, it is also sensitive to changes in appearance of each component caused by occlusion, pose, and variations in expression. In contrast, component-level features focus on the description of a single facial component, and thus are independent of changes in appearance of the other components. In this way the information generated by these two feature levels is complementary and appropriate fusion of the two promotes robustness to interference.

\subsection{Implementation Details}
Similar to most state-of-the-art face representation schemes, e.g.,~\cite{chen2013blessing,taigman2014deepface},
MDML-DCPs also benefits from recent progress in face alignment algorithms that accurately locate dense facial feature points in real-time.
The MDML-DCPs face representation scheme is shown in Fig.~\ref{fig:MDMLDcpsFramework}.
In MDML-DCPs, a high performance face alignment algorithm based on~\cite{xiong2013supervised} is first employed to locate 49 facial feature points
(as shown in Fig.~\ref{fig:MDMLDcpsSixComponent}a), before applying the two geometric rectifications.

The similarity transformation is based on the two detected eye centers. In the rectified image, we compute MD-DCPs at two holistic levels: 1) MD-DCPs of non-overlapping regions of the external cropped face image (including facial contour), and 2) MD-DCPs of non-overlapping regions of the internal cropped face image (without facial contour). The first feature encodes both facial contour and facial components while the second feature focuses on encoding facial components only, which are free from background interference. For clarity, we denote the two features as MDML-DCPs-H1 and MDML-DCPs-H2.

The affine transformation is determined by the three detected facial feature points: the centers of the two eyes and the center of the mouth. In the rectified image, one holistic-level feature denoted as MDML-DCPs-H3, and six component-level features referred to MDML-DCPs-C1 to MDML-DCPs-C6, are computed based on the detected dense facial feature points. As shown in Fig.~\ref{fig:MDMLDcpsSixComponent}b, the method for feature extraction around each facial feature point is similar to the approaches used in~\cite{prince2008tied,chen2013blessing}:
centered on each facial feature point, a patch of size $M\times M$ pixels is located and further divided into $J\times J$ non-overlapping regions. The concatenated MD-DCPs
feature of the $J^{2}$ regions forms the description of the feature point.

MDML-DCPs-H3 and MDML-DCPs-C1 to C6 are formed by respectively concatenating the descriptions of different facial feature
points. As shown in Fig.~\ref{fig:MDMLDcpsSixComponent}, MDML-DCPs-H3 selects 21 facial feature points over all facial components, while MDML-DCPs-C1 to C6 select feature points on only one particular facial component. Elements of the three holistic-level
features and six component-level features are normalized by the square root. Together, the set of the nine normalized feature vectors form the MDML-DCPs face representation.

\section{Face Recognition Algorithm}
In this section, the face matching problem is addressed using the proposed MDML-DCPs face representation scheme.
First, one classifier is built for each of the nine feature vectors.
Then, the similarity scores of the nine classifiers are fused by linear SVM or simple averaging.

Two algorithms are considered: Whitened Principal Component Analysis (WPCA; an unsupervised learning algorithm)
and Probabilistic Linear Discriminant Analysis (PLDA; a supervised learning algorithm)~\cite{li2012probabilistic,prince2007probabilistic}.
The choice of which of the two algorithms to use is dataset dependent:
for datasets that have a training set with multiple face images for each subject, we choose PLDA; otherwise, WPCA is used.
For more recent high-performance classifiers, readers can refer to~\cite{chen2012bayesian,liu2015Classification}.

\subsection{WPCA}
Principal Component Analysis (PCA) learns an orthogonal projection matrix $U$ from training data and projects high-dimensional feature vector $x$ to the low-dimensional vector $y$,
\begin{equation}
y = U^Tx.
\end{equation}

The columns of $U$ are composed of the leading eigenvectors of the covariance matrix of the training data.
However, the first few eigenvectors in $U$ encode mostly variations in illumination and expression, rather than information that discriminates identity.
The whitening transformation tackles this problem by normalizing the contribution of each principal component
\begin{equation}
y = {\left( {U{\Lambda ^{ - 1/2}}} \right)^T}x,
\end{equation}
where $\Lambda  = diag\left\{ {\lambda _1,\lambda _2, \cdots } \right\}$ with ${\lambda _i}$ being the $i$th leading eigenvalue.
After projecting the facial feature vectors to the low-dimensional subspace using WPCA,
the similarity score between two feature vectors ${y_1}$ and ${y_2}$ is measured by the cosine metric
\begin{equation}
Sim\left( {{y_1},{y_2}} \right) = \frac{{y_1^T{y_2}}}{{\left\| {{y_1}} \right\|\left\| {{y_2}} \right\|}}.
\end{equation}

\subsection{PCA combined with PLDA}
The feature vectors in MDML-DCPs are high-dimensional. To effectively apply PLDA, the dimensionality of the nine feature vectors is first reduced by PCA.

PLDA models the face data generation process
\begin{equation}
{x_{ij}} = \mu  + F{h_i} + G{w_{ij}} + {\varepsilon _{ij}},
\end{equation}
where $x_{ij}$ denotes the $j$th face data of the $i$th individual. $\mu$ is the mean of all face data.
$F$ and $G$ are factor matrices whose columns are the basis vectors of the between-individual subspace and the within-individual subspace, respectively.
$h_i$ is the latent identity variable that is constant for all images of the $i$th subject.
$w_{ij}$ and $\varepsilon _{ij}$ are noise terms explaining intra-personal variance.

It is shown in~\cite{li2012probabilistic} that the identification and verification problems can be solved
by computing the log-likelihood ratio that whether two observed images share the same identity variable $h$ or not.
In this paper, we refer to the log-likelihood ratio as similarity score for consistency with the case of the WPCA classifier.

\section{EXPERIMENTS}
\begin{figure*}
\centering
\subfigure[]{
    \label{fig:FeretPealFrgc:a}
    \begin{minipage}[c]{0.75\textwidth}
    \centering
    \includegraphics[width=1.00\linewidth,height=0.4286\linewidth]{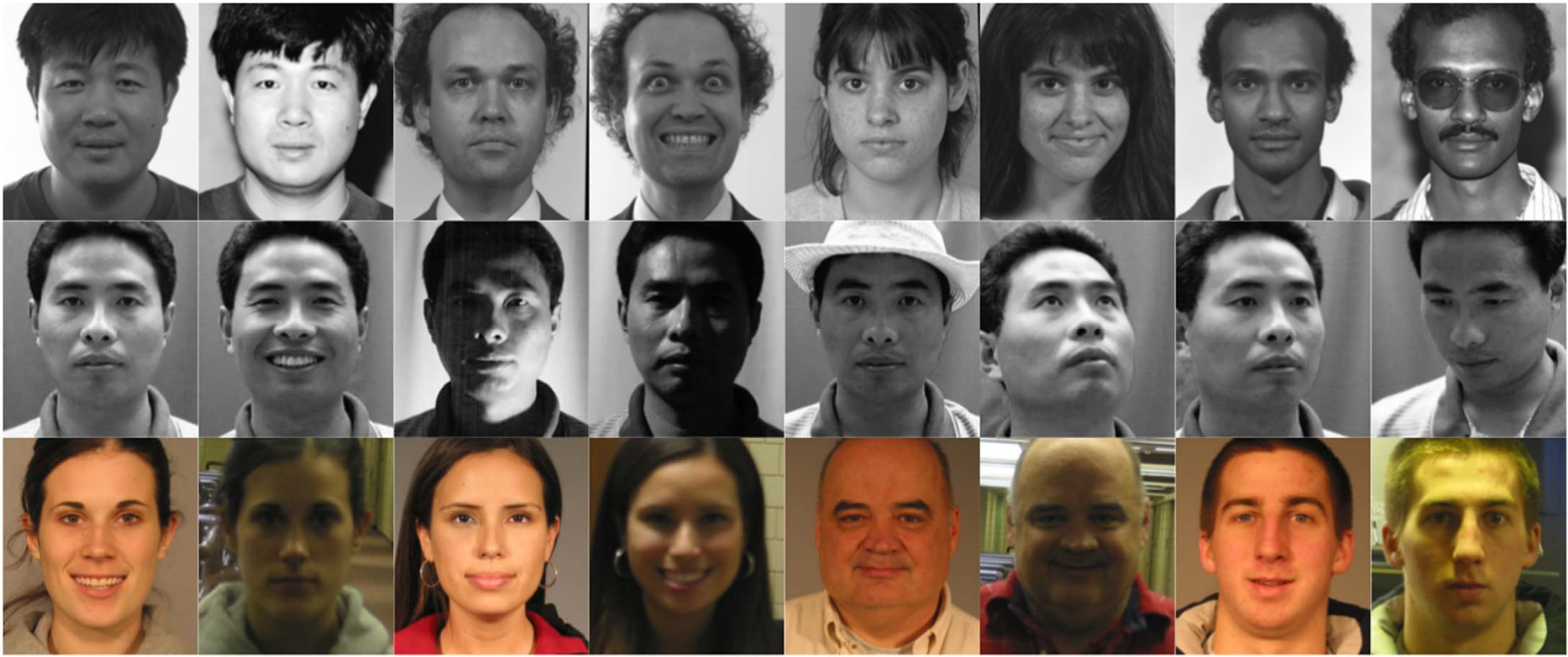}
    \end{minipage}%
    }%
\subfigure[]{
    \label{fig:FeretPealFrgc:b}
    \begin{minipage}[c]{0.25\textwidth}
    \centering
    \includegraphics[width=0.8571\linewidth,height=1.2857\linewidth]{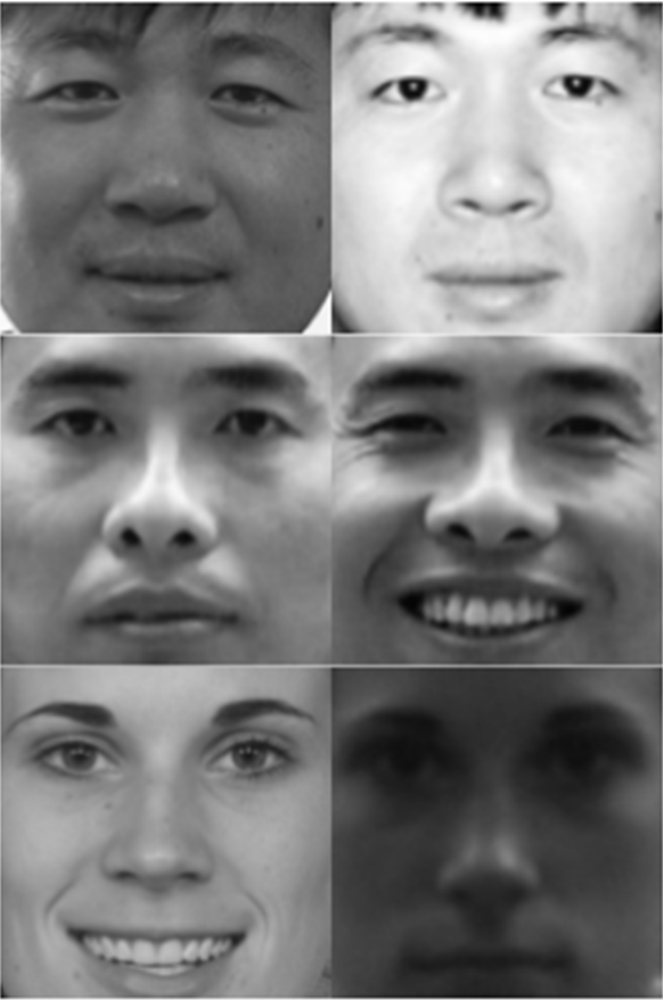}\\
    \end{minipage}}
\caption{(a) Sample images from FERET (first row), CAS-PEAL-R1 (second row) and FRGC 2.0 (third row) containing typical variations in each database.
(b) Samples of normalized images for experiments from Section 5.1 to 5.3.}
\label{fig:FeretPealFrgc}
\end{figure*}

\begin{figure}
\centering
\includegraphics[width=3.2in]{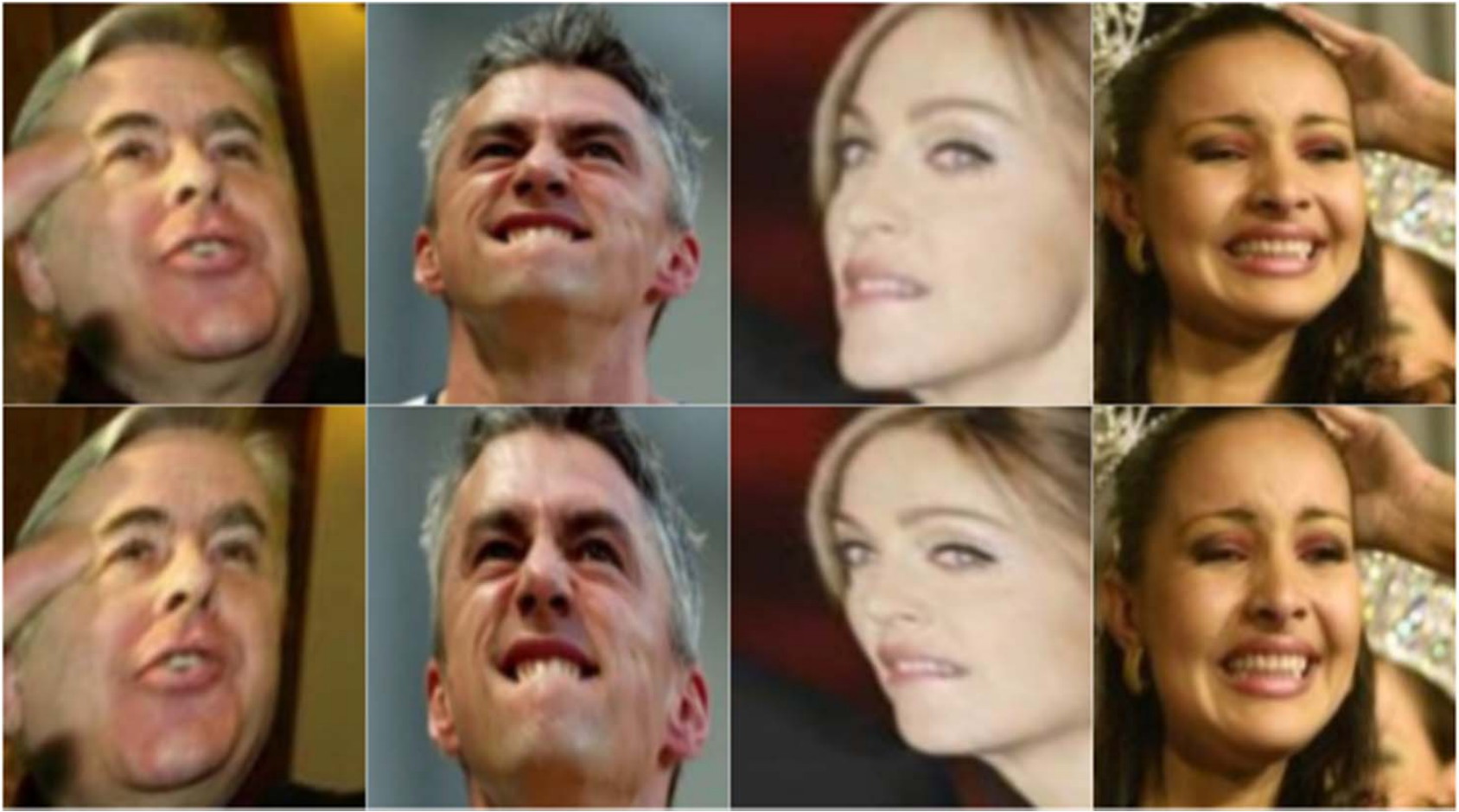}
\caption{Sample images from LFW. Images in the two rows are aligned by a similarity transformation and an affine transformation, respectively.}
\label{fig:LfwIm}
\end{figure}

In this section, the proposed DCP and MDML-DCPs are extensively evaluated in both face identification and face verification tasks.
Experiments are conducted on four publicly available large-scale face databases: FERET, CAS-PEAL-R1, FRGC 2.0, and LFW.
Example images of the four databases are shown in Figs.~\ref{fig:FeretPealFrgc} and~\ref{fig:LfwIm}.

The FERET database contains one gallery set Fa and four probe sets, i.e., Fb, Fc, Dup1, and Dup2.
In this paper, the standard face identification protocol specified in~\cite{phillips2000feret} is employed.

The CAS-PEAL-R1 database includes one training set, one gallery set, and nine probe sets.
Each of the nine probe sets is restricted to one type of variations.
In detail, the PE, PA, PL, PT, PB, and PS probe sets correspond to variations in expression, accessory, lighting, time, background, and distance of frontal faces, respectively.
The PU, PM, and PD probe sets correspond to one type of pose variation.
The standard identification protocol defined in~\cite{gao2008cas} is followed.

The FRGC 2.0 database incorporates data for six face verification experiments~\cite{phillips2005overview}.
We focus on Experiments 1 and 4.
The shared target set of Experiments 1 and 4 is collected from frontal facial images taken under controlled illumination,
while images in their query sets are captured under controlled and uncontrolled conditions, respectively.
Verification rates at 0.1\% FAR are reported.

The LFW database~\cite{LFWTech} contains 13,233 unconstrained face images that are organized into two ``Views''.
View 1 is for model selection and parameter tuning while View 2 is for performance reporting.
Two paradigms are used to exploit the training set in View 2, an image-restricted paradigm and an image-unrestricted paradigm.
In the first paradigm, only the officially defined image pairs are available for training.
In the second paradigm, the identity information of the training images can be used.
We report the mean verification accuracy and standard error of the mean ($S_E$) on the View 2 data.

Four experiments are conducted.
First, the dual-cross grouping mode for DCP is empirically proved to achieve the maximum joint Shannon entropy.
Second, the performance of DCP is compared with eleven state-of-the-art face image descriptors.
Third, the performance of MD-DCPs is evaluated to determine the contribution of multi-directional filtering by FDG.
Finally, the power of the MDML-DCPs face representation scheme is presented.
More experimental results, e.g., parameter selection of DCP, are available in the appendix.

In the first three experiments, we focus on the evaluation of face image descriptors.
In these three experiments, all face images are cropped and resized to $128\times128$ ($rows\times columns$)
pixels with the eye centers located at $(34, 31)$ and $(34, 98)$, as shown in Fig.~\ref{fig:FeretPealFrgc}b.
For LFW, we use the aligned version (LFW-A)~\cite{wolf2011effective}, while for the other three databases, images are cropped according to the eye coordinates provided by the databases.
The cropped images are photometrically normalized using a simple operator (denoted as TT) developed by Tan \& Triggs~\cite{tan2010enhanced}
and then encoded by one face image descriptor.
Each encoded face image is divided into $N\times N$ non-overlapping regions.
Concatenating the code histograms of these regions forms the representation for the image.

In the fourth experiment, we aim to demonstrate that the proposed MDML-DCPs face representation scheme has excellent performance characteristics.
In this experiment, all face images are normalized by two geometric transformations, as illustrated in Fig.~\ref{fig:LfwIm}.
In both transformations, face images are resampled to $180\times162$ pixels with the eye centers mapped to $(66, 59)$ and $(66, 103)$.
For the affine-transformed face images, the centers of the mouths are unified to $(116, 81)$.
TT is applied to the resampled images for photometric normalization.

\subsection{Empirical Justification for Dual-Cross Grouping}
Section 2.3 intuitively suggests that the dual-cross grouping mode is optimal from the perspective of the joint Shannon entropy maximization.
In this experiment, we empirically validate this point on the FERET database.
For each of the 35 possible grouping modes, $DC{P_i}\left( {0 \le i \le 7} \right)$ are divided into two subsets.
Then, the joint Shannon entropy for each of the two subsets are calculated on one image using~\eqref{eq:jointEntropy} and are summed together.
The above process is repeated on the 1,196 gallery images of the FERET database. The mean value of the summed joint Shannon entropy is recorded.

\begin{figure}
\centering
\includegraphics[width=0.72\linewidth]{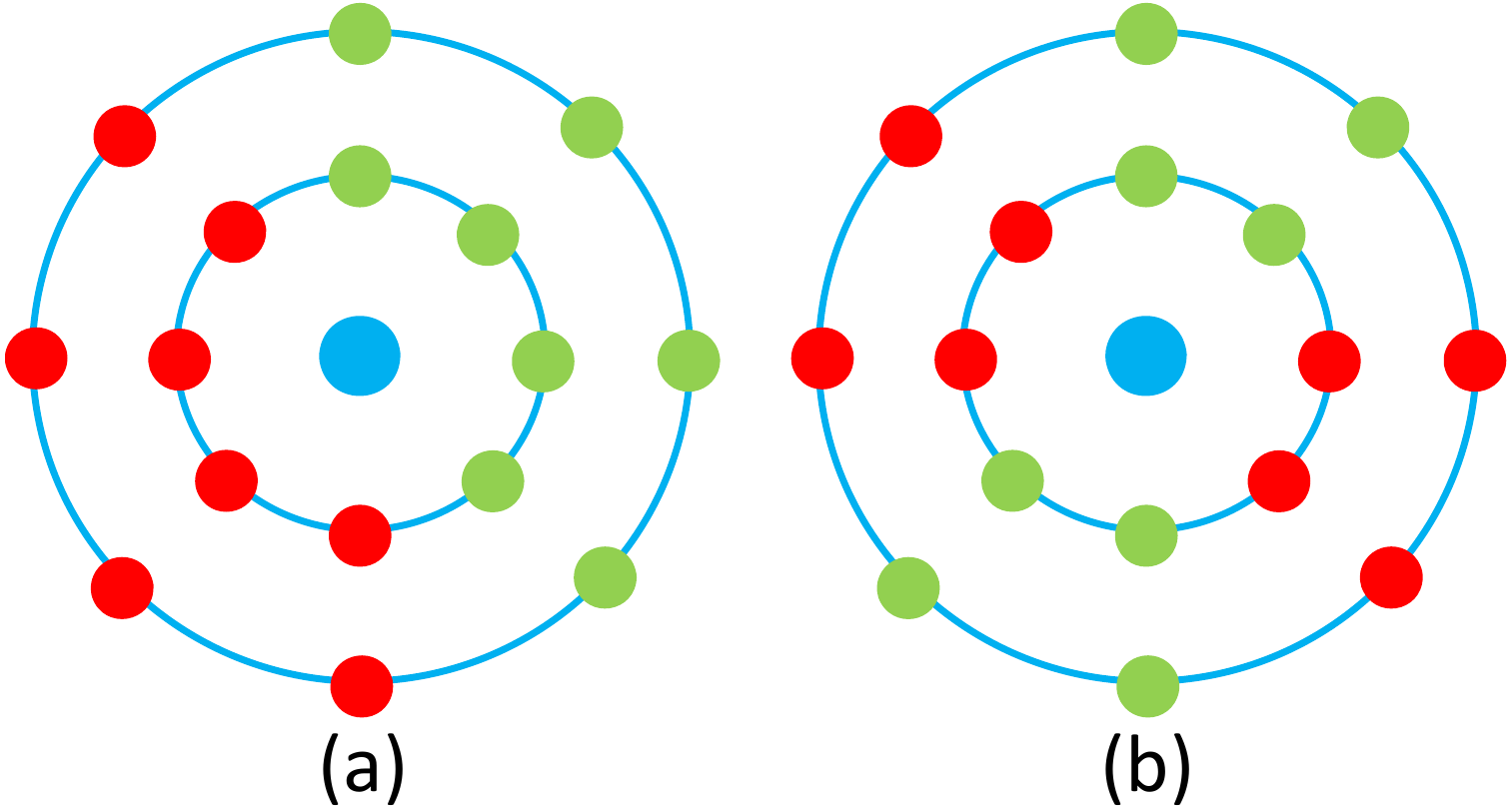}
\caption{Another two representative grouping modes for the eight sampling directions of DCP.
Sampled points of the same colour belong to the same subset.}
\label{fig:OtherGroupingModes}
\end{figure}

\begin{figure}
\centering
\includegraphics[width=0.9\linewidth]{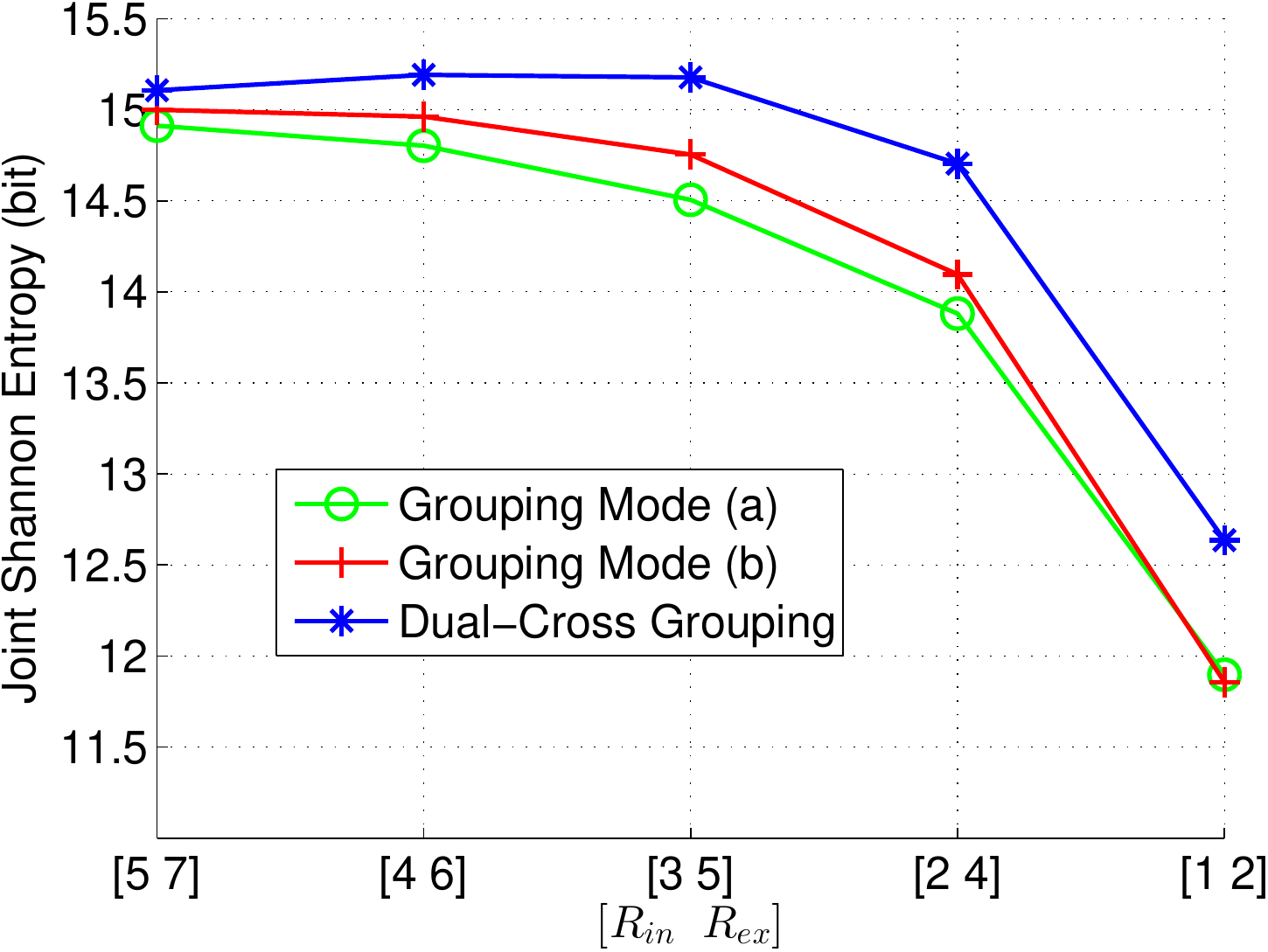}
\caption{Joint Shannon entropy as a function of $R_{in}$ and $R_{ex}$. Three grouping modes are evaluated in this figure:
modes (a) and (b) in Fig.~\ref{fig:OtherGroupingModes} and the dual-cross grouping.}
\label{fig:GroupingModeEntropy}
\end{figure}

Experimental results show that dual-cross grouping mode achieves the highest joint Shannon entropy among all 35 grouping modes.
Fig.~\ref{fig:GroupingModeEntropy} characterizes the joint Shannon entropy as a function of $R_{in}$ and $R_{ex}$
and illustrate the superiority of dual-cross grouping mode by comparing it with two representative grouping modes,
exampled in Fig.~\ref{fig:OtherGroupingModes}.
Note that the joint Shannon entropy is related to the sampling radii of DCP:
smaller sampling radii mean stronger dependence among the sampled points, which results in a smaller joint Shannon entropy.
The dual-cross grouping mode achieves the highest entropy under all sets of the radii values. Therefore, the dual-cross grouping mode is empirically optimal.

\subsection{Evaluation of the Performance of DCP}
In this section, the performance of DCP, DCP-1, and DCP-2 for both face identification and face verification tasks are evaluated.
To better illustrate the advantages of DCP, the performance of eleven state-of-the-art face image descriptors,
i.e., LBP~\cite{ahonen2006face}, LTP~\cite{tan2010enhanced}, LPQ~\cite{ahonen2008recognition}, POEM~\cite{vu2012enhanced},
Local Gabor XOR Patterns (LGXP)~\cite{xie2010fusing}, Multi-scale LBP (MsLBP), Multi-scale tLBP (MsTLBP)~\cite{trefny2010extended}, Multi-scale dLBP (MsDLBP)~\cite{trefny2010extended},
LQP~\cite{hussain2012face}, DFD~\cite{lei2014learning}, and CHG~\cite{choi2012complementary} are also presented.
\footnote{For Table~\ref{tab:DcpFERET} to Table~\ref{tab:DcpLFW}, a superscript $^*$ means that the results are cited from the original papers.
A suffix `-Flip' means that the descriptor adopts the `flip' trick~\cite{hussain2012face}.}

The first eight descriptors are hand-crafted and they are easy to implement. We therefore test them together with DCP.
It is worth noting that the cropped face image data for these descriptors are exactly the same.
All of them extract features from photometrically normalized images by TT.
The parameters for each of the descriptors and TT are carefully tuned on each database and different distance metrics (chi-squared or histogram intersection) are tested.
Finally, the best results for each descriptor are reported.
Therefore, the experimental results directly compare the recognition capabilities of the descriptors.

On the other hand, LQP, DFD, and CHG are learning-based descriptors,
which are complicated to implement. In this paper, their performance on FERET and LFW is directly cited from the original papers, even though the experimental settings are different
from that of DCP. For LQP, we cite its performance with the image filtering step by Gabor wavelets, which is more robust to illumination variation.
Before presenting the detailed experimental results, the feature size (number of histogram bins) for each descriptor except CHG
(CHG is not a histogram-based descriptor) is listed in Table~\ref{tab:featureSize}.
In the following experiments, LBP-based descriptors are implemented without using uniform coding~\cite{ojala2002multiresolution}
(using uniform coding actually degrades the performance of LBP, LTP, and POEM, et al.).
For POEM, we compute LBP codes on four-orientation gradient magnitude maps, so its feature size is 1024 in this paper.

\begin{table}[!t]
\renewcommand{\arraystretch}{1.3}
\caption{Feature Size of the Investigated Face Image Descriptors}
\label{tab:featureSize}
\centering
\begin{tabular}{|l|c||l|c|}
\hline
Descriptor                            &Feature Size             &Descriptor         &Feature Size\\ \hline\hline
LBP~\cite{ahonen2006face}             &256                      &MsDLBP~\cite{trefny2010extended}      &512\\ \hline
MsLBP~\cite{ojala2002multiresolution} &512                      &LQP~\cite{hussain2012face}            &300\\ \hline
LTP~\cite{tan2010enhanced}            &512                      &DFD~\cite{lei2014learning}            &1024\\ \hline
LPQ~\cite{ahonen2008recognition}      &256                      &DCP-1                                 &256\\ \hline
POEM~\cite{vu2012enhanced}            &1024                     &DCP-2                                 &256\\ \hline
LGXP~\cite{xie2010fusing}             &640                      &DCP                                   &512\\ \hline
MsTLBP~\cite{trefny2010extended}      &512                      &                                      &\\ \hline
\end{tabular}
\end{table}

\subsubsection{Face Identification: FERET}
Face identification experiments are conducted on the FERET and CAS-PEAL-R1 databases. The rank-1 identification rates on the four probe sets of FERET are presented in Table~\ref{tab:DcpFERET}. We make four observations:

\begin{enumerate}[\setlabelwidth{12}]
\item While high performance is achieved by all descriptors on the well controlled Fb and Fc probe sets, DCP still outperforms MsLBP by over 1\%.
      In particular, DCP achieves a perfect identification rate on the Fc set.
\item There is a substantial performance drop for all descriptors on Dup1 and Dup2 probe sets,
      in which images contain moderate expression and illumination variations.
      DCP performs best on both sets, with a margin of 1.28\% on Dup2 by comparing with the second best descriptor.
      The performance of MsTLBP ranks second on Dup1 and LGXP ranks second on Dup2.
      It is worth pointing out that LGXP depends on 80 expensive convolutions and produces a larger feature size,
      while both DCP and MsTLBP have low computational complexity.
\item Both DCP-1 and DCP-2 perform better than most of the other descriptors at lower computational cost. Their time and memory costs are exactly the same as those of LBP,
      suggesting the sampling and encoding strategies of DCP are highly effective.
\item As expected, the performance of DCP is better than both DCP-1 and DCP-2, which means that DCP-1 and DCP-2 contain complementary information. When DCP-1 and DCP-2 are combined, the mean identification rate increases by over 1\%.
\end{enumerate}

\begin{table}[!t]
\renewcommand{\arraystretch}{1.3}
\caption{Identification Rates for Different Descriptors on FERET}
\label{tab:DcpFERET}
\centering
\begin{tabular}{|l|c|c|c|c|c|}
\hline
                    &Fb     &Fc     &Dup1   &Dup2   &Mean\\ \hline\hline
LBP                 &96.90  &98.45  &83.93  &82.48  &90.44\\ \hline
LTP                 &96.90  &98.97  &83.93  &83.76  &90.89\\ \hline
LPQ                 &97.41  &99.48  &82.69  &81.62  &90.30\\ \hline
POEM                &98.24  &99.48  &82.83  &82.05  &90.65\\ \hline
LGXP                &97.32  &99.48  &85.46  &85.47  &91.93\\ \hline
MsLBP               &97.07  &98.97  &83.38  &83.33  &90.69\\ \hline
MsTLBP              &98.16  &99.48  &85.73  &85.04  &92.10\\ \hline
MsDLBP              &95.65  &97.94  &79.09  &79.49  &88.04\\ \hline
LQP-Flip$^*$        &\textbf{99.50} &99.50  &81.20  &79.90  &90.03\\ \hline
DFD$^*$             &99.20  &98.50  &85.00  &82.90  &91.40\\ \hline
CHG$^*$             &97.50  &98.50  &85.60  &84.60  &91.55\\ \hline \hline
DCP-1               &97.91  &98.45  &84.49  &84.19  &91.26\\ \hline
DCP-2               &97.99  &99.48  &84.35  &85.04  &91.72\\ \hline
\textbf{DCP}        &98.16  &\textbf{100.0}  &\textbf{86.29}  &\textbf{86.75}  &\textbf{92.80}\\ \hline
\end{tabular}
\end{table}

\subsubsection{Face Identification: CAS-PEAL-R1}
Identification results on the nine probe sets of CAS-PEAL-R1 are shown in Table~\ref{tab:DcpPEAL}.
For each descriptor, results are reported with the set of parameters that achieves the highest mean identification rate over all nine probe sets.
We make the following observations:

\begin{enumerate}[\setlabelwidth{12}]
\item In general, the performance of all the descriptors is good on the PE, PA, PT, PB, and PS probe sets,
      with DCP producing the highest mean identification rate and MsTLBP the second highest.
      The performance of all descriptors is poor on the remaining PL, PU, PM, and PD probe sets due to serious illumination and pose variations.
\item DCP outperforms MsLBP and LTP by 2.5\% and 3.08\% on the PL set, respectively. However, the performance of DCP is lower than LPQ, POEM, and LGXP.
      The PL images contain not only rich illumination variation,
      but also serious image blur, which explains the superior performance of the blur-invariant descriptor LPQ.
      LGXP and POEM benefit from image filtering steps that turn the gray-scale image into gradient images.
      We show in Section 5.3 that by introducing multi-directional gradient filtering by FDG, the performance of DCP is significantly improved.
\item DCP exhibits excellent robustness to pose variations on the PU, PM, and PD probe sets, with superior mean recognition rate by as much as 3.44\%.
      This result is encouraging, since pose variation is a major challenge in unconstrained face recognition~\cite{ding2015comprehensive}.
\end{enumerate}

\begin{table*}[!t]
\renewcommand{\arraystretch}{1.3}
\caption{Rank-1 Identification Rates for Different Face Image Descriptors on the Nine Probe Sets of PEAL}
\label{tab:DcpPEAL}
\centering
\begin{tabular}{|l|c|c|c|c|c|c|c|c|c|c|c|c|}
\hline
           &PE     &PA     &PT     &PB      &PS     &PL     &PU     &PM     &PD     &\tabincell{c}{Mean\\(PE-PS)}   &\tabincell{c}{Mean\\(PE-PL)}   &\tabincell{c}{Mean\\(PU-PD)}\\ \hline\hline
LBP        &94.27  &91.82  &100.0  &\textbf{99.46}  &\textbf{99.64}  &46.90  &60.32  &83.66  &44.60   &97.04  &88.68   &62.86\\ \hline
LTP        &94.39  &91.77  &100.0  &\textbf{99.46}  &\textbf{99.64}  &47.17  &61.32  &84.46  &44.68   &97.05  &88.74   &63.49\\ \hline 
LPQ        &93.95  &92.39  &100.0  &99.28  	&99.27  &57.16	         &57.78  &86.46  &44.76  &96.98   &90.34  &63.00\\ \hline
POEM       &95.54  &92.39  &100.0  &\textbf{99.46}  &\textbf{99.64}  &54.66  &58.14  &85.12  &42.52   &97.41  &90.28   &61.93\\ \hline 
LGXP       &94.97  &91.33  &100.0  &99.28   &\textbf{99.64}  &\textbf{63.26} &39.46  &66.83  &22.91   &97.04  &\textbf{91.41}  &43.07\\ \hline
MsLBP      &95.16  &92.04  &100.0  &\textbf{99.46}  &\textbf{99.64}  &47.75  &61.76  &85.16  &44.88   &97.26  &89.01   &63.93\\ \hline
MsTLBP     &95.41  &92.74  &100.0  &\textbf{99.46}  &\textbf{99.64}  &48.06  &60.98  &87.34  &45.48   &97.45      &89.22     &64.60\\ \hline
MsDLBP     &92.42  &90.63  &100.0  &99.28   &99.27  &48.11  &55.00   &82.03  &37.03  &96.32  &88.29   &58.02\\ \hline
DFD$^*$    &\textbf{99.30} &\textbf{94.40} &-       &-      &-    &59.00  &-    &-    &-       &-      &-     &-\\ \hline\hline
DCP-1      &95.99  &92.60  &100.0  &98.73   &99.27  &46.37   &60.68  &85.74  &48.14   &97.32   &88.83  &64.85\\ \hline
DCP-2      &95.54  &91.90  &100.0  &98.92   &\textbf{99.64}  &43.91  &60.64  &83.08   &43.62   &97.20  &88.32   &62.45\\ \hline
DCP        &96.11  &92.82  &100.0  &99.10   &\textbf{99.64}  &50.25  &\textbf{65.39}  &\textbf{87.44}  &\textbf{51.30}  &\textbf{97.53}  &89.65  &\textbf{68.04}\\ \hline 
\end{tabular}
\end{table*}

\subsubsection{Face Verification: FRGC 2.0}
\begin{table}[!t]
\renewcommand{\arraystretch}{1.3}
\caption{Verification Results on the FRGC 2.0 Experiment 1}
\label{tab:DcpFRGCExp1}
\centering
\begin{tabular}{|l|c|c|c|c|}
\hline
           &ROC I     &ROC II     &ROC III  &Mean\\ \hline\hline
LBP        &89.93     &87.53      &84.83    &87.43\\ \hline
LTP        &90.98     &88.63      &86.07    &88.56\\ \hline
LPQ        &89.96     &87.38      &84.48    &87.27\\ \hline
POEM       &90.60     &88.26      &85.63    &88.16\\ \hline
LGXP       &91.89     &88.80      &85.33    &88.67\\ \hline
MsLBP      &90.78     &88.45      &85.87    &88.37\\ \hline
MsTLBP     &\textbf{93.24}     &91.16      &88.83    &91.08\\ \hline
MsDLBP     &87.65     &84.93      &81.96    &84.85\\ \hline\hline
DCP-1      &91.40     &89.14      &86.65    &89.06\\ \hline
DCP-2      &92.96     &90.78      &88.42    &90.72\\ \hline
\textbf{DCP}  &93.22  &\textbf{91.21}      &\textbf{88.93}    &\textbf{91.12}\\ \hline
\end{tabular}
\end{table}

\begin{table}[!t]
\renewcommand{\arraystretch}{1.3}
\caption{Verification Results on the FRGC 2.0 Experiment 4}
\label{tab:DcpFRGCExp4}
\centering
\begin{tabular}{|l|c|c|c|c|}
\hline
                &ROC I     &ROC II     &ROC III  &Mean\\ \hline\hline
LBP             &18.62     &19.12      &20.07    &19.27\\ \hline
LTP             &20.21     &21.39      &22.99    &21.53\\ \hline
LPQ             &22.53     &23.25      &24.39    &23.39\\ \hline
POEM            &29.99     &30.60      &\textbf{31.46}    &30.68\\ \hline
\textbf{LGXP}   &\textbf{32.44}  &\textbf{31.81} &31.13  &\textbf{31.79}\\ \hline
MsLBP           &19.34     &19.80      &20.81    &19.98\\ \hline
MsTLBP          &20.04     &20.25      &21.02    &20.44\\ \hline
MsDLBP          &18.34     &18.85      &19.60    &18.93\\ \hline\hline
DCP-1           &20.10     &20.57      &21.51    &20.73\\ \hline
DCP-2           &18.22     &18.74      &19.58    &18.85\\ \hline
DCP             &21.78     &22.49      &23.59    &22.62\\ \hline 
\end{tabular}
\end{table}

Face verification experiments are conducted on the FRGC 2.0 and LFW databases.
Verification rates at 0.1\% FAR on Experiments 1 and 4 of FRGC 2.0 are listed in
Tables~\ref{tab:DcpFRGCExp1} and~\ref{tab:DcpFRGCExp4}, respectively.
As mentioned above, the query sets of the two experiments are composed of controlled and uncontrolled images, respectively.
The query images in Experiment 4 are degraded by serious image blur and significant illumination variation, making this dataset very challenging.
We make the following observations:

\begin{enumerate}[\setlabelwidth{12}]
\item In Experiment 1, DCP shows the best performance. Both DCP-1 and DCP-2 achieve better performance than most of the existing descriptors.
      These observations are consistent with the results on FERET.
\item In Experiment 4, the mean verification rate of DCP is higher than MsLBP and LTP by 2.64\% and 1.09\%, respectively.
      Due to the lack of an image filtering step in DCP, its performance is lower than that of LGXP and POEM.
      This result is consistent with that seen for the PL set of CAS-PEAL-R1, whose images also feature serious illumination variation and image blur.
\end{enumerate}

\subsubsection{Face Verification: LFW}
LFW is a very challenging dataset since its images contain large pose, expression, and illumination variations.
Experiments in this section follow the image-restricted paradigm.
For each descriptor, its parameters are tuned on the View 1 data, and performance on the View 2 data is reported in Table~\ref{tab:DcpLFW}. We observe that:

\begin{enumerate}[\setlabelwidth{12}]
\item Among the manually-designed descriptors, DCP achieves the best performance while DCP-1 ranks second.
      This result is consistent with the results on FERET and Experiment 1 on FRGC 2.0.
      The robustness of DCP to pose variations is consistent with the results on CAS-PEAL-R1.
\item LGXP does not perform well on LFW because LFW images contain significant pose variations.
      Although LGXP is robust to serious illumination variations, it is sensitive to pose variations,
      consistent with the results seen from experimentation on the non-frontal probe sets of CAS-PEAL-R1.
\item Compared with the two learning-based descriptors, the performance of DCP is better than that of LQP, but is worse than that of DFD.
      However, DCP has clear advantage in both time and memory costs.
      Besides, DFD adopts supervised learning algorithms to enhance its discriminative power,
      while DCP is independent of any learning algorithms.
      We show in Section 5.4 that by applying supervised learning algorithms to the extracted DCP feature, superior performance can be achieved.
\end{enumerate}

\begin{table}[!t]
\renewcommand{\arraystretch}{1.3}
\caption{Mean Verification Accuracy on the LFW View 2 Data}
\label{tab:DcpLFW}
\centering
\begin{tabular}{|l|c||l|c|}
\hline
                       &Accuracy(\%)$\pm S_E$         &                       &Accuracy(\%)$\pm S_E$  \\ \hline\hline
LBP                    &72.43 $\pm$ 0.49              &MsDLBP                 &72.17 $\pm$ 0.59\\\hline
LTP                    &72.65 $\pm$ 0.52              &LQP-Flip$^*$           &75.30 $\pm$ 0.26\\\hline
LPQ                    &72.68 $\pm$ 0.46              &\textbf{DFD-Flip$^*$}  &\textbf{80.02 $\pm$ 0.50}\\\hline
POEM                   &73.98 $\pm$ 0.56              &DCP-1                  &74.50 $\pm$ 0.48\\ \hline
LGXP                   &70.58 $\pm$ 0.43              &DCP-2	              &73.28 $\pm$ 0.48\\ \hline
MsLBP                  &72.88 $\pm$ 0.50              &DCP                    &75.00 $\pm$ 0.64\\ \hline
MsTLBP                 &74.12 $\pm$ 0.57              &DCP-Flip               &76.37 $\pm$ 0.71\\ \hline
\end{tabular}
\end{table}

\subsubsection{Discussion}
\begin{figure}
\centering
\includegraphics[width=0.95\linewidth]{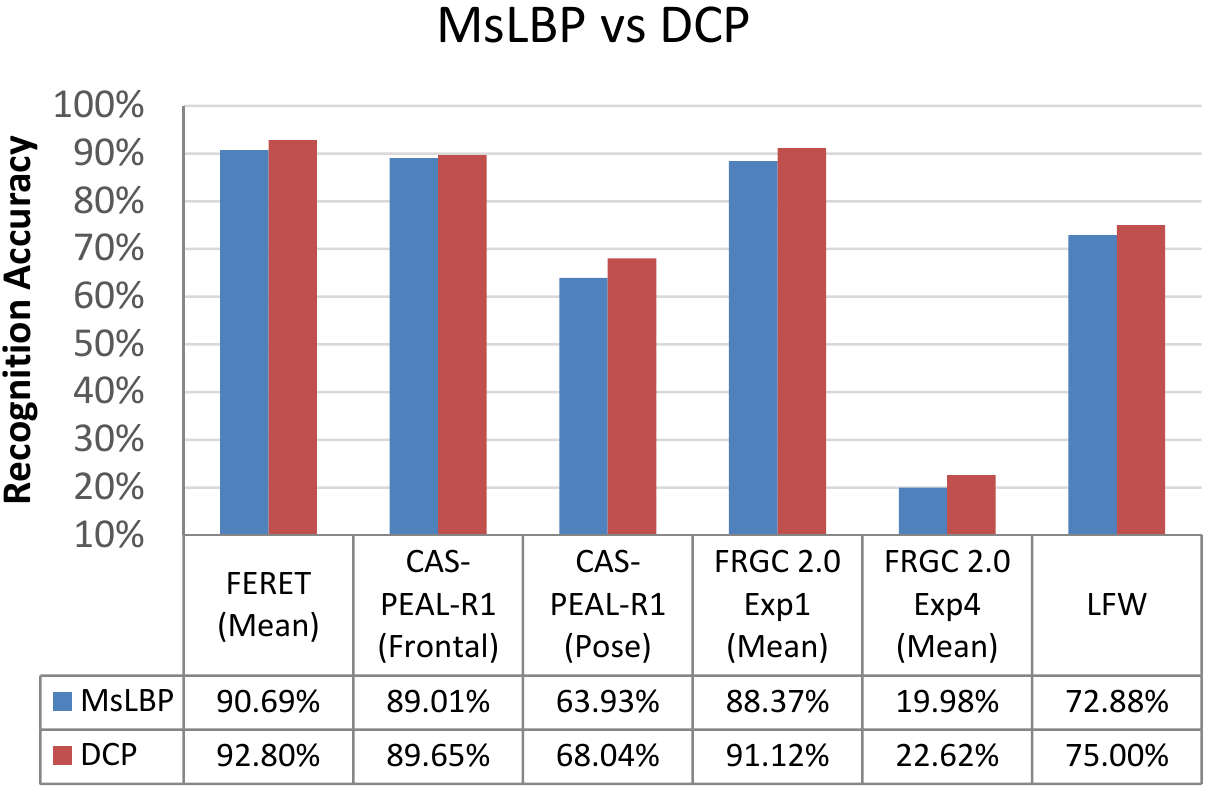}
\caption{Performance comparison between DCP and MsLBP on the four face datasets.}
\label{fig:MsLBPvsDCP}
\end{figure}

The experimental results in Section 5.2 reveal that DCP performs quite well across a range of evaluations.
In particular, DCP significantly outperforms MsLBP, which is of exactly the same time and memory costs as DCP, as shown in Fig.~\ref{fig:MsLBPvsDCP}.
The performance of MsTLBP is usually slightly worse than DCP on well-controlled datasets.
However, DCP considerably outperforms MsTLBP on the challenging PU and PD probe sets of CAS-PEAL-R1, and the Dup2 probe set of FERET.
This indicates DCP is a more robust descriptor to pose and aging factors.

The excellent performance of DCP can be explained as follows.
It has large sampling size to encode more discriminative information;
it encodes the second-order statistics in the most informative directions on human face;
and the dual-cross grouping strategy ensures good complementarity between the divided two encoders.

\subsection{The Contribution of Multi-directional Filtering}
In Section 5.2, DCP is shown to have strong discriminative power and excellent robustness to expression, pose, and moderate illumination variations.
It has also been shown that when there is serious
illumination variation and image blur, the performance of DCP is degraded due to the lack of an image filtering step.
In this section, we show that MD-DCPs that extracts the DCP feature on FDG-filtered images is robust to illumination variation.
The roles of FDG are two-fold: first, it converts the face image into gradient maps, which are more robust to illumination variation,
and second, FDG is proven to achieve high SNR~\cite{canny1986computational}, which is vital for the performance of face image descriptors for low quality images.

Throughout this paper, the number of filtering orientations by FDG is fixed to 4 ($\theta = 0, \pi/4, \pi/2,$ and $3\pi/4$) and the parameter $\sigma$ of the Gaussian kernel is set at 1.0. Experiments are conducted on three datasets:

\begin{itemize}[\setlabelwidth{12}]
\item A new FERET probe set is constructed by collecting the 146 most challenging FERET probe images.
      The FERET Fa set is still utilized as the gallery set. The identification rate of DCP on this probe set is only 17.12\%.
      In contrast, the identification rate when MD-DCPs is used increases to 33.56\%.
\item For the PL probe set of CAS-PEAL-R1, the identification rates of DCP and MD-DCPs are 50.25\% and 65.23\%, respectively.
      When considered along with the previously best results shown in Table~\ref{tab:DcpPEAL},
      it is evident that the performance of MD-DCPs is superior to that of LPQ (57.16\%), POEM (54.66\%), and LGXP (63.26\%).
\item For the face verification task on FRGC 2.0 Experiment 4, the mean verification rate of MD-DCPs at 0.1\% FAR for ROC I, ROC II, and ROC III is 30.74\%.
      Together with the previously best results shown in Table~\ref{tab:DcpFRGCExp4}, it is clear that MD-DCPs outperforms LPQ (23.39\%) and POEM (30.68\%).
      LGXP outperforms MD-DCPs with the mean verification rate of 31.79\%, but at the cost of 80 expensive convolution operations.
\end{itemize}

In conclusion, MD-DCPs consistently achieves excellent performance,
which means that multi-directional filtering by FDG is indeed helpful for removing illumination variation and enhancing the robustness of face image descriptors on low SNR images.

\subsection{Performance Evaluation of MDML-DCPs}
Before presenting the performance of MDML-DCPs, we first unify most of its parameters for testing the four databases.
For the holistic-level feature MDML-DCPs-H1, we compute it within the area whose top left corner is located at $(33, 27)$,
while the bottom right corner is located at $(154, 136)$ in the normalized image by similarity transformation.
The holistic-level feature MDML-DCPs-H2 is computed within the area where the two corners are located at $(36, 41)$ and $(140, 122)$, respectively.
Both areas are divided into $9\times 9$ non-overlapping regions.
For MDML-DCPs-H3 and MDML-DCPs-C1 to C6, we consistently use a patch size $M\times M$ of $40\times 40$ and the number of regions $J\times J$ within the patch to be $4\times 4$.

The two parameters $\sigma_1$ and $\sigma_2$ for TT are set to 1.4 and 2.0 on FERET, CAS-PEAL-R1, and FGRC 2.0.
Since TT in fact slightly degrades the performance of MDML-DCPs on the LFW View 1 data,
the photometric normalization step is omitted for the experiment on LFW.
The subspace dimensions of WPCA and PCA for all nine features are set to 600.
The dimensions of both the between-individual subspace and within-individual subspace of PLDA are set to 100.
There are only two remaining sets of parameters to tune over the four databases: the first is the sampling radii ${R_{in}}$ and ${R_{ex}}$ of DCP,
and the other is the cost parameter $c$ of the linear SVM~\cite{CC01a}.
The optimal values of ${R_{in}}$ and ${R_{ex}}$ for FERET, CAS-PEAL-R1, and FGRC 2.0 are ${R_{in}=2}$ and ${R_{ex}=3}$.
$c$ is set at ${10^{ - 4}}$ for FRGC 2.0.
For LFW, ${R_{in}}$, ${R_{ex}}$, and $c$ are set to 4, 6, and 1.0, respectively, based on the results on the View 1 data.

\subsubsection{Face Identification: FERET}
As there is no officially defined training set for FERET, we test MDML-DCPs with the WPCA-based classifiers.
The nine WPCA projection matrices are trained on the 1,196 FERET gallery images.
For simplicity, the similarity scores computed by the nine classifiers are fused by averaging with equal weights.
Performance comparison between MDML-DCPs with the state-of-the-art approaches is shown in Table~\ref{tab:MDMLDcpsFeret}.
The first two approaches use supervised learning algorithms.
The others utilize the unsupervised learning algorithm WPCA.
The performance of MDML-DCPs is superior even to those that employ a supervised learning algorithm.
In particular, MDML-DCPs achieves the best results on the Fc, Dup1, and Dup2 probe sets.
On the Fb set, three images are misclassified by MDML-DCPs, two of which are due to the mislabeling of subjects in the database itself.

\begin{table}[!t]
\renewcommand{\arraystretch}{1.3}
\caption{Identification Rates for Different Methods on FERET}
\label{tab:MDMLDcpsFeret}
\centering
\begin{tabular}{|l|c|c|c|c|}
\hline
                                                        &Fb     &Fc     &Dup1   &Dup2\\ \hline\hline
AMFG07` (L,G) + KDCV~\cite{tan2007fusing}               &98.00  &98.00  &90.00  &85.00\\ \hline
TIP10` LGBP + LGXP + LDA~\cite{xie2010fusing}           &99.00  &99.00  &94.00  &93.00\\ \hline
BMVC12` G-LQP + WPCA~\cite{hussain2012face}             &\textbf{99.90}  &\textbf{100.0}  &93.20  &91.00\\ \hline
PAMI14` DFD + WPCA~\cite{lei2014learning}           &99.40  &\textbf{100.0}  &91.80  &92.30\\ \hline
\textbf{MDML-DCPs + WPCA}                               &99.75  &\textbf{100.0}  &\textbf{96.12}  &\textbf{95.73}\\ \hline
\end{tabular}
\end{table}

\subsubsection{Face Identification: CAS-PEAL-R1}
For experiments on CAS-PEAL-R1, we present MDML-DCPs results with both WPCA and PLDA, with DCP set as the baseline.
Both WPCA and PLDA are trained on all 1,200 training images.
In both approaches, the similarity scores are fused by simple averaging without weighting.
A comparison between the performance of MDML-DCPs with other state-of-the-art methods is presented in Table~\ref{tab:MDMLDcpsPeal}.
We make the following observations:

\begin{table*}[!t]
\renewcommand{\arraystretch}{1.3}
\caption{Rank-1 Identification Rates for Different Methods on the Nine Probe Sets of PEAL}
\label{tab:MDMLDcpsPeal}
\centering
\begin{tabular}{|l|c|c|c|c|c|c|c|c|c|}
\hline
                                                &PE     &PA     &PL     &PT     &PB      &PS     &PU     &PM     &PD\\ \hline\hline
TIP07` LGBPHS~\cite{zhang2005local,zhang2007histogram}         &95.20  &86.80  &51.00  &100.0  &98.70   &98.90  &-     &-      &-\\ \hline
TIP07` HGPP~\cite{zhang2007histogram}           &96.80  &92.50  &62.90  &98.40  &99.80   &99.60  &-      &-      &-\\ \hline
SP09` Weighted LLGP~\cite{xie2009learned}       &98.00  &92.00  &55.00  &-      &-       &-      &-      &-      &-\\ \hline
PAMI14` DFD + WPCA~\cite{lei2014learning}   &99.00  &96.90  &63.90  &-      &-       &-      &-      &-      &-\\ \hline
DCP + $\rchi^2$                                 &96.11  &92.82  &50.25  &100.0  &99.10   &99.64  &65.39  &87.44  &51.30\\ \hline
MDML-DCPs + PLDA                                &98.22  &97.51  &63.26  &100.0  &\textbf{100.0}   &100.0  &34.31  &70.24  &33.59\\ \hline
\textbf{MDML-DCPs + WPCA}                       &\textbf{99.62}  &\textbf{99.21}  &\textbf{82.92}  &\textbf{100.0}  &99.82   &\textbf{100.0}  &\textbf{68.29}  &\textbf{98.10}  &\textbf{53.92}\\ \hline
\end{tabular}
\end{table*}

\begin{enumerate}[\setlabelwidth{12}]
\item ``MDML-DCPs + WPCA'' achieves the best performance for eight of the nine probe sets of CAS-PEAL-R1.
      In particular, it outperforms DCP by 32.67\% on the challenging PL probe set.
\item Compared with DCP, there is limited performance promotion made by ``MDML-DCPs + WPCA'' on the PU and PD sets.
      The reason is that the training set contains only frontal images while the PU and PD images possess large pose variations.
      Therefore the two probe sets benefit little from training.
\item The identification rates by ``MDML-DCPs + PLDA'' are lower than ``MDML-DCPs + WPCA'',
      which suggests that there might be a gap between the distribution of training and testing data.
      In particular, the discriminative subspace learnt by PLDA on the frontal images does not generalize to the three non-frontal probe sets.
\end{enumerate}

\subsubsection{Face Verification: FRGC 2.0}
The performance of MDML-DCPs on Experiments 1 and 4 of the FRGC 2.0 database is discussed.
The FRGC 2.0 database provides a large training set, on which PLDA and linear SVM are trained.
For each subject in the training set, 12 images are randomly selected and there are 2,664 images in total for SVM training,
for which 13,320 intra-personal images pairs and 13,320 inter-personal image pairs are generated according to the method described in~\cite{LFWTech}.
All the remaining images are used to train the PLDA model.
The performance of the proposed algorithm, together with other state-of-the-art approaches, is shown in Table~\ref{tab:MDMLDcpsFRGC}.
We observe that:

\begin{table*}[!t]
\renewcommand{\arraystretch}{1.3}
\caption{Verification Rates at 0.1\% FAR for Different Methods on the FRGC 2.0 Experiments 1 and 4}
\label{tab:MDMLDcpsFRGC}
\centering
\begin{tabular}{|c|l|c|c|c||c|c|c|}
\hline
\multirow{2}{*}{} & \multirow{2}{*}{} &
\multicolumn{3}{c||}{Experiment 1} &
\multicolumn{3}{c|}{Experiment 4} \\
\cline{3-8}
  & & ROC I & ROC II & ROC III & ROC I & ROC II & ROC III \\
\hline\hline

\multirow{6}{*}{\rotatebox{90}{Single Descriptor}}
& TIP09` Gabor + LDA~\cite{su2009hierarchical}                              &-      &-      & 97.00 &-       &-      & 83.00 \\
\cline{2-8}
& TIP10` Gabor + Kernel LDA~\cite{tan2010enhanced}                          &-      &-      &-      &-       &-      & 80.00 \\
\cline{2-8}
& PAMI13` Multiscale LPQ (MLPQ) + Kernel fusion~\cite{chan2012multiscale}   & 98.99 & 98.68 & 98.37 & 89.26  & 89.80 & 90.36 \\
\cline{2-8}
& PAMI13` Multiscale LBP (MLBP) + Kernel fusion~\cite{chan2012multiscale}   & 98.79 & 98.50 & 98.21 & 86.77  & 87.54 & 88.21 \\
\cline{2-8}
& MDML-DCPs + PLDA + Score averaging                                        & 99.59 & 99.41 & 99.22 & 93.40  & 93.18 & 92.89 \\
\cline{2-8}
& \textbf{MDML-DCPs + PLDA + Linear SVM}                                    & \textbf{99.61} & \textbf{99.43} & \textbf{99.25} & \textbf{93.91}  & \textbf{93.64} & \textbf{93.39} \\\hline\hline
\multirow{7}{*}{\rotatebox{90}{Multiple Descriptors}}
& AMFG07` Gabor + LBP + KDCV~\cite{tan2007fusing}                           &-      &-      &-      &-       &-      & 83.60 \\
\cline{2-8}
& TIP09` Gabor + DCT + LDA~\cite{su2009hierarchical}                        &-      &-      & 98.00 &-       &-      & 89.00 \\
\cline{2-8}
& ICB09` Hybrid RCrQ with Gabor + MLBP + DCT~\cite{liu2009robust}           &-      &-      &-      &-       &-      & 92.43 \\
\cline{2-8}
& TIP10` Gabor + LBP + Kernel LDA~\cite{tan2010enhanced}                    &-      &-      &-      &-       &-      & 88.10 \\
\cline{2-8}
& TIP10` LGBP + LGXP + Block-based LDA~\cite{xie2010fusing}                 & 98.60 & 98.00 & 97.30 & 83.90  & 84.70 & 85.20 \\
\cline{2-8}
& TIP11` Hybrid Fourier feature + LDA~\cite{hwang2011face}                  & 96.38 & 95.11 & 93.90 & 81.82  & 81.50 & 81.14 \\
\cline{2-8}
& PAMI13` MLPQ + MLBP + Kernel fusion~\cite{chan2012multiscale}             & 99.04 & 98.77 & 98.49 & 90.30  & 90.94 & 91.59 \\\hline
\end{tabular}
\end{table*}

\begin{enumerate}[\setlabelwidth{12}]
\item The MDML-DCPs approach achieves superior performance on both Experiment 1 and 4 with the single descriptor DCP.
      Score fusion by linear SVM also works slightly better than score averaging.
\item On Experiment 1, the performance of MDML-DCPs is nearly perfect. On Experiment 4, it outperforms the current state-of-the-art method~\cite{liu2009robust} by 0.9\%. Moreover, MDML-DCPs utilizes only a single face image descriptor, while the method in~\cite{liu2009robust} relies on color information and makes use of three descriptors: LBP, Gabor, and Fourier features.
\end{enumerate}

Different from the approaches in Table~\ref{tab:MDMLDcpsFRGC},
it is shown in~\cite{li2013fusing} that by conducting normalization on the similarity matrices which contain similarity scores of each pair of target and query images,
verification rates can be improved significantly.
However, this operation may not be suitable for the general face verification problems, where only a pair of target and query images is available each time.
Therefore, we provide results of MDML-DCPs without score normalization in this paper.

\subsubsection{Face Verification: LFW}
Experiments in this section follow the image-unrestricted paradigm of the LFW database.
Both the PLDA classifier and the more recently proposed Joint Bayesian (JB) classifier~\cite{chen2012bayesian} are tested for face matching,
and both linear SVM and score averaging are tested for the score fusion.
For each of the 10-fold cross validations, 
the remaining nine subsets are partitioned into two: the first eight subsets are used for the training of PCA, PLDA, and JB models,
while the last subset is used to train the linear SVM or learn the optimal threshold of similarity scores fused by averaging.
According to the description in~\cite{LFWTech}, 2,500 intra-personal image pairs and 2,500 inter-personal image pairs are generated using the images in the last subset.
No outside training data are employed.
In addition, following~\cite{wolf2011effective,li2012probabilistic},
all images with right profile faces are flipped to the left profile faces.

\begin{figure}
\centering
\includegraphics[width=0.95\linewidth,height=0.78\linewidth]{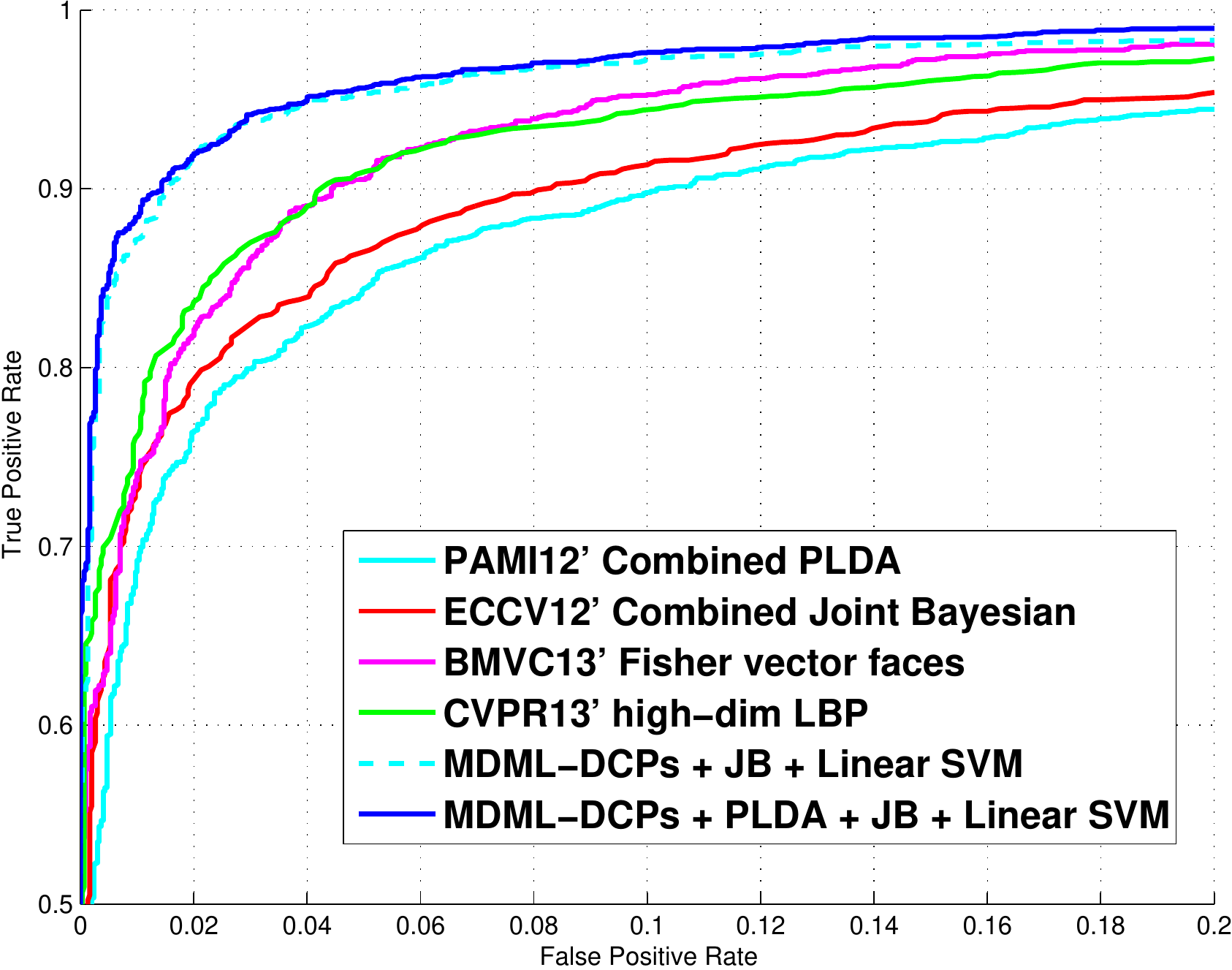}
\caption{ROC curves of the MDML-DCPs method and other state-of-the-art methods in the unrestricted paradigm.}
\label{fig:LfwRocMdmlDcps}
\end{figure}

A comparison of the classification accuracies is presented in Table~\ref{tab:MDMLDcpsLFW}.
All the methods in Table~\ref{tab:MDMLDcpsLFW} follow the image-unrestricted training paradigm and do not use outside training data.
The ROC curves are plotted in Fig.~\ref{fig:LfwRocMdmlDcps}.
We make the following observations:

\begin{enumerate}[\setlabelwidth{12}]
\item MDML-DCPs outperforms the current state-of-the-art method~\cite{chen2013blessing} by over 2.2\%.
      It is worth noting that~\cite{chen2013blessing} also employs dense facial landmarks for face representation.
      MDML-DCPs outperforms~\cite{li2012probabilistic}, which also employs the PLDA-based classifier, by over 5\%.
      The above facts suggest that the MDML-DCPs face representation scheme is more effective than previous approaches.
\item Score fusion by linear SVM results in better performance than using score averaging.
      Moreover, with either linear SVM or score averaging, MDML-DCPs is able to achieve better performance than existing methods,
      and using both fusion methods the standard error of classification accuracy is consistently smaller than the other approaches.
\item The JB-based classifiers slightly outperform PLDA-based classifiers.
      By fusing the 18 similarity scores produced by the two kinds of classifiers with linear SVM,
      we achieve a better mean verification rate of 95.58\%.
\end{enumerate}

\begin{table}[!t]
\renewcommand{\arraystretch}{1.3}
\caption{Mean Verification Accuracy on the LFW View 2 Data}
\label{tab:MDMLDcpsLFW}
\centering
\begin{tabular}{|l|c|c|}
\hline
                                                        &Accuracy(\%) $\pm S_E$\\ \hline\hline
PAMI12` Combined PLDA~\cite{li2012probabilistic}        &90.07 $\pm$ 0.51\\\hline
ECCV12` Combined Joint Bayesian~\cite{chen2012bayesian} &90.90 $\pm$ 1.48\\\hline
ICCV13` VMRS~\cite{barkanfast}                          &92.05 $\pm$ 0.45\\\hline
BMVC13` Fisher vector faces~\cite{simonyan2013fisher}   &93.03 $\pm$ 1.05\\\hline
CVPR13` high-dim LBP + JB~\cite{chen2013blessing}       &93.18 $\pm$ 1.07\\\hline
\textbf{MDML-DCPs + PLDA + Score averaging}             &\textbf{94.57 $\pm$ 0.30}\\\hline
\textbf{MDML-DCPs + PLDA + Linear SVM}                  &\textbf{95.13 $\pm$ 0.33}\\\hline
\textbf{MDML-DCPs + JB + Linear SVM}                    &\textbf{95.40 $\pm$ 0.33}\\\hline
\textbf{MDML-DCPs + PLDA + JB + Linear SVM}             &\textbf{95.58 $\pm$ 0.34}\\\hline
\end{tabular}
\end{table}

\section{Conclusion}
Due to the degradation of face image quality and large variations of illumination, pose, and expression, the recognition of unconstrained face images is a challenging task.
Solving this problem demands work in at least two areas: development of an effective face image descriptor and a comprehensive face representation scheme.
To achieve this goal, we make the following contributions:

\begin{enumerate}[\setlabelwidth{12}]
\item We presented a novel face image descriptor named Dual-Cross Patterns.
      DCP encodes second-order statistics in the most informative directions within a face image.
      Experimentation on four large-scale face databases shows that DCP has superior discriminative power and is robust to pose, expression,
      and moderate variations in illumination.
\item A comprehensive and systematic comparison of fourteen state-of-the-art face image descriptors is conducted on the four face databases.
      Detailed analysis is provided.
      Conclusions about the advantages and disadvantages of the descriptors can be drawn from these results.
\item The proposed MDML-DCPs face representation scheme comprehensively incorporates both holistic-level DCP features and component-level DCP features.
      Exploiting the single descriptor DCP, MDML-DCPs consistently achieves the best results on the four databases.
      In particular, MDML-DCPs improves the verification rate of ROC III in Experiment 4 of FRGC 2.0 to 93.39\%
      and outperforms the state-of-the-art result on LFW by 2.4\%.
\end{enumerate}

This work helps to expedite the design of practical face image descriptors and face representation schemes.


%

\appendices
\section{Parameter Selection of DCP}
In this experiment, we take the FERET database~\cite{phillips2000feret} for example to illustrate the influence of the DCP parameters on its performance.
The face images are normalized to $128\times 128$ pixels, as described in Section 5 of the paper.
There are three parameters in DCP, the radii ${R_{in}}$ and ${R_{ex}}$ (with different values of ${R_{in}}$ and ${R_{ex}}$
capturing information on different scales) and the region number $N$.
A larger value of $N$ helps to preserve spatial information but makes the descriptor more sensitive to misalignment errors.
The chi-squared metric is used to measure the similarity of two face images.
In this experiment, LBP is also evaluated for comparison with a sampled points number of eight and exploiting all 256 LBP codes.
The radius of LBP is restricted to the value of ${R_{in}}$ to highlight the importance of including more sampling points for effective pattern encoding.

\begin{figure}
\centering
\includegraphics[width=1.0\linewidth]{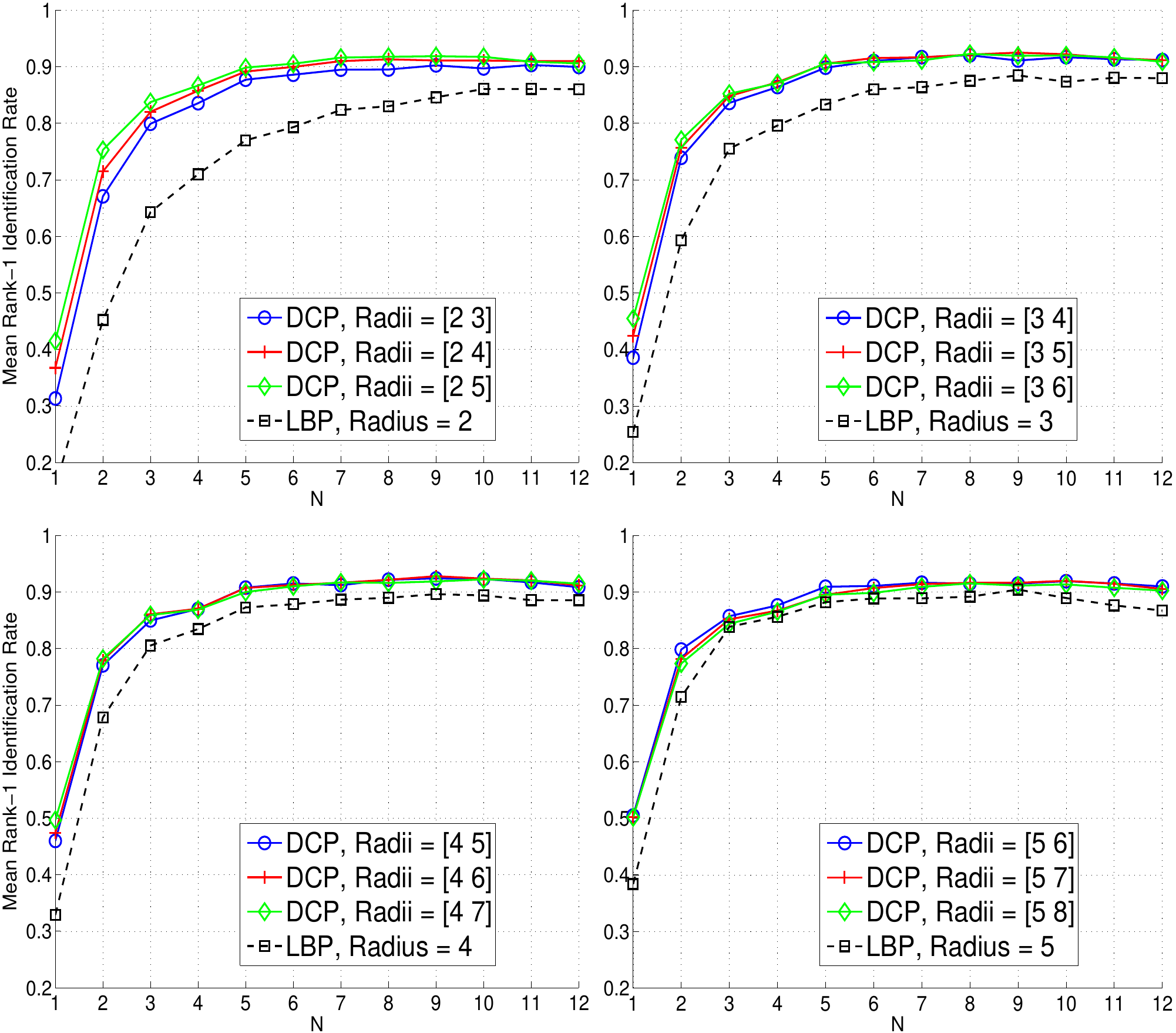}
\caption{The mean rank-1 identification rates of DCP and LBP on four FERET probe sets as a function of $N$.}
\label{fig:FeretDcpParaTuning}
\end{figure}

The mean rank-1 identification rates against $N$ for the four FERET probe sets using DCP and LBP are plotted in Fig.~\ref{fig:FeretDcpParaTuning}.
Optimal performance of DCP is achieved when $N=9$, ${R_{in}=4}$ and ${R_{ex}=6}$.
DCP consistently achieves better performance than LBP with all sets of parameters.
In particular, DCP outperforms LBP by 10\% to 30\% when $N\le2$,
which justifies the motivation of DCP to include more sampling points for pattern encoding.

The chi-squared metric has been used to measure the similarity of two face images, but it is noted that DCP works equally well when assessed using the histogram intersection metric.
The mean identification rates using DCP measured with the chi-squared and histogram intersection metrics are 92.80\% and 92.66\%, respectively.

\section{Performance of DCP on Facial Aging Database}
We have also tested the performance of the face image descriptors on the Morph database~\cite{ricanek2006morph}.
The face images in Morph are geometrically and photometrically normalized as described in Section 5 of the paper.
We follow the protocol defined in~\cite{akhtar2013face} and report the performance of the face image descriptors in Table~\ref{tab:DcpMorph}.
The same as~\cite{akhtar2013face}, a subset of 1,700 images in the Morph database is randomly selected for experiment.
It is shown that the proposed DCP descriptor achieves the best performance on the Morph database, indicating that DCP is also robust to the aging factor.

\begin{table}[!t]
\renewcommand{\arraystretch}{1.3}
\caption{AUC Values of Face Image Descriptors on the Morph Database}
\label{tab:DcpMorph}
\centering
\begin{tabular}{|l|c||l|c|}
\hline
                       &Mean$\pm$ Std                 &                       &Mean$\pm$ Std  \\ \hline\hline
LBP                    &0.9479 $\pm$ 0.0045           &MsTLBP                 &0.9506 $\pm$ 0.0037\\\hline
LTP                    &0.9499 $\pm$ 0.0048           &MsDLBP                 &0.9393 $\pm$ 0.0044\\\hline
LPQ                    &0.9454 $\pm$ 0.0044           &DCP-1                  &0.9499 $\pm$ 0.0035\\ \hline
POEM                   &0.9500 $\pm$ 0.0045           &DCP-2	              &0.9529 $\pm$ 0.0034\\ \hline
LGXP                   &0.9369 $\pm$ 0.0040           &\textbf{DCP}           &\textbf{0.9557 $\pm$ 0.0044}\\ \hline
MsLBP                  &0.9494 $\pm$ 0.0037           &                       &    \\ \hline
\end{tabular}
\end{table}

\ifCLASSOPTIONcompsoc
  \section*{Acknowledgments}
\else
  \section*{Acknowledgment}
\fi
This work is partially supported by Australian Research Council Projects DP-140102164 and FT-130101457.
C. Ding is partially supported by China Scholarship Council.
J. Choi and L. S. Davis are partially supported by NSF Grant IIS1262121.

\ifCLASSOPTIONcaptionsoff
  \newpage
\fi



\bibliographystyle{IEEEtran}





\end{document}